\documentclass{article}
\usepackage[preprint]{neurips_2026}
\makeatletter
\renewcommand{\@notice}{}
\makeatother
\usepackage[utf8]{inputenc}
\usepackage[T1]{fontenc}
\usepackage{hyperref}
\usepackage{url}
\usepackage{textcomp}
\usepackage{booktabs}
\usepackage{amsmath}
\usepackage{amsthm}
\usepackage{amssymb}
\usepackage{graphicx}
\usepackage{microtype}
\usepackage{xcolor}
\usepackage{multirow}
\usepackage{array}
\usepackage{colortbl}
\usepackage{caption}

\definecolor{schema1col}{RGB}{20,55,105}     
\definecolor{warmcol}{RGB}{190,80,45}        
\definecolor{tealcol}{RGB}{42,115,95}        
\definecolor{purpcol}{RGB}{105,75,148}       
\usepackage{float}
\usepackage{pgfplots}
\usepackage{tikz}
\pgfplotsset{compat=1.18}
\usetikzlibrary{positioning, arrows.meta, shapes.geometric, fit, backgrounds}

\newtheorem{definition}{Definition}

\usepackage{titlesec}
\titlespacing*{\section}{0pt}{2ex plus 0.5ex minus 0.2ex}{1ex plus 0.2ex}
\titlespacing*{\paragraph}{0pt}{1ex plus 0.3ex minus 0.2ex}{0.5em}
\title{Data Language Models: A New Foundation Model Class for Tabular Data}

\author{
  Eda Erol \quad Giuliano Pezzoli \quad Özer Cem Kelahmet \\
  SchemaLabs \\
  \texttt{research@schemalabs.ai}
}

\begin{document}
\maketitle

\begin{abstract}
Every major data modality now has a foundation model that understands it natively:
text has language models, images have vision models, audio has audio models.
Tabular data, the modality on which many consequential real-world AI decisions
are made, does not.
Every approach to tabular AI today, from gradient-boosted trees to the latest
tabular foundation models, requires a preprocessing pipeline before any model
can consume the data.
None of them understand tabular data as a modality.
We introduce the \textbf{Data Language Model} (DLM), the missing foundation model
for tabular data.
A DLM understands tables the way a language model understands sentences: natively,
without serialization or preprocessing, directly from raw cell values.
It is the tabular data layer on which AI models, agents, and vertical AI
applications can be built, eliminating the preprocessing pipelines that currently
stand between raw data and every AI system that consumes it.
We present \textbf{Schema-1}, the first DLM: a 140M parameter model trained on
more than 2.3M synthetic and real-world tabular datasets.
Schema-1 outperforms gradient-boosted ensembles, AutoML stacks, and the tabular
foundation models we evaluate on established row-level prediction benchmarks.
On missing value reconstruction it achieves lower reconstruction error than
all classical statistical methods and frontier large language models on mean
performance across conditions, establishing that structural understanding of a
dataset's own distributional geometry is more useful for imputation than world
knowledge encoded in language.
It identifies the industry sector of any unseen dataset from raw cell values alone,
reliably across any domain, a task no prior tabular model can perform.
It is the native tabular understanding layer that has been missing from the AI
stack.
\end{abstract}

\section{Introduction}
\label{sec:introduction}

\paragraph{The centrality of tabular data.}
Structured tables are the medium through which machine learning determines the most
consequential outcomes in deployed systems: the dosing adjustment for a patient,
the fraud flag on a transaction, the failure prediction for a component before its
next scheduled inspection, the significance threshold in a clinical trial.
These are tabular prediction problems.
Their inputs are rows and columns; their outputs are derived from the statistical
patterns in cell values.
Electronic health records, trading books, loan portfolios, risk ledgers, financial
transaction logs, industrial sensor archives, experimental measurement datasets, and
regulatory filings are all tabular.
Structured tables are the dominant working data format across machine
learning practice~\cite{borisov2022deep}, and the medium of record in
healthcare, finance, manufacturing, and regulatory compliance across every
major industry~\cite{vanbreugel2024tabular,shwartziv2022tabular}.
The same structured data now forms the primary substrate of a new wave of enterprise
AI: vertical applications purpose-built for specific industries, and AI agents that
reason over enterprise data to take autonomous actions.
In both paradigms, tabular data is not a secondary input: it is the primary
information layer through which enterprise AI perceives the world and acts on it.
The models making these decisions are almost uniformly tree-based: gradient-boosted
ensembles that have remained the empirical state of the art on structured data
benchmarks for more than ten years.
The reason is well established.
Tabular data has three properties that vision and language architectures cannot
exploit: heterogeneous column types, irregular marginal distributions of cell
values, and no spatial or sequential regularity.
Gradient-boosted trees have inductive biases well matched to this setting; deep
architectures historically have not~\cite{grinsztajn2022tree,shwartziv2022tabular}.
Language and vision have absorbed the majority of foundation model research
investment.
Tabular data, the substrate on which the majority of consequential enterprise
decisions are made, has not.
This constitutes a systematic misallocation: despite tabular data's dominance in
deployed machine learning systems and its growing centrality to vertical and agentic
AI, the architectural principles required for genuine cross-domain generalization on
structured inputs remain incompletely understood, and the design of tabular foundation
models remains an active area of research~\cite{vanbreugel2024tabular}.
The present work identifies and addresses a capability that no prior approach has
been designed to provide: understanding what datasets are, their domain, their
relational structure, and the semantic content of their features, not only predicting
from them.

\paragraph{The gap in current approaches.}
Three research directions address machine learning on tabular data, each effective
for the prediction and data science workflows it was designed for.
Tree-based methods and their AutoML descendants are powerful and mature tools:
given labeled data, they learn conditional distributions reliably and scale to
production deployments across a broad range of structured prediction tasks.
Tabular foundation models have taken a meaningful step further, demonstrating that
a single pretrained model can generalize to unseen classification and regression
tasks without task-specific training, reducing the cold-start cost of building
prediction systems~\cite{hollmann2023tabpfn}.
Numerical tabular foundation models such as TabPFN~\cite{hollmann2023tabpfn} are
designed to treat all features identically, a deliberate choice for generalization
across datasets with arbitrary column configurations: the column named
\texttt{patient\_glucose} and the column named \texttt{feature\_3} produce the same
internal representation, because the architecture is not designed to incorporate
column semantics or domain identity.
Semantics-aware variants such as ConTextTab~\cite{spinaci2025contexttab} do
encode column names, bringing richer input representations to structured
prediction; but that semantic signal is channeled entirely into the numerical
output and does not appear on the output side: the prediction carries no
account of what the input dataset represents.
A third direction applies large language models to tables via
serialization~\cite{hegselmann2023tabllm,su2024tablegpt}, embedding semantic
reasoning into data workflows by representing rows and columns as natural language.
Each direction serves well the data science and prediction workflows it was built
for.
The paradigm they were not built for is the one now emerging: vertical AI
applications that operate autonomously on raw enterprise data across any domain,
and AI agents that reason over structured data to take actions without
human-assembled pipelines in the loop.
Both require a model that knows what data it is looking at, not only what to predict
from it.

Domain identification is not a step that can be added after the fact.
In every current tabular pipeline, the engineer who assembles the system
examines the data, decides what it represents, selects the target column, and
chooses the preprocessing and modeling configuration appropriate for that
domain.
An agent receiving a clinical record and a cross-border payment log in the same
workflow cannot apply appropriate reasoning to either without knowing which is which,
and today that knowledge must be supplied externally, as configuration, before
inference can begin.
For a single labeled dataset in a known domain, this is a routine setup cost.
For an autonomous system that must operate across any enterprise data it encounters
at runtime, this requirement is a structural barrier: no pipeline can be assembled
in advance for data sources the system has not yet encountered, because domain
context must be specified by a human for each new source.
Domain identification embedded inside the model removes this barrier entirely.

The gap that persists across all three directions is not in prediction capability.
It is in the understanding layer that vertical and agentic AI require as their
foundation.
No existing approach was designed to identify what a dataset is from its own
structural and distributional properties, without external labels, metadata, or a
hand-assembled pipeline.
A model that cannot determine what domain it is operating in, cannot represent what
its features mean without being told, and requires a preprocessing pipeline before
it can function is not a viable base layer for autonomous enterprise AI: it is a
component that still requires humans to do the understanding.
Tabular data is the only major data modality without a foundation model that
provides this layer.
Text, image, and audio each have model classes built specifically for their
modality: models that consume the raw signal directly, learn its structure, and
serve as the foundation on which applications and agents are built.
Tabular data has no such layer.
No training objective has ever required a model to understand what a table is from
its own distributional and structural properties: domain identification has always
been handled upstream, by the humans assembling the pipeline.
The DLM paradigm places that capacity inside the model.

\paragraph{This work.}
We introduce the \textbf{Data Language Model} (DLM) as the paradigm that closes
this gap.
A DLM understands tables natively: it processes the semantic content of column
identifiers alongside per-column distributional summaries, raw cell values, and
missing value patterns, deriving a joint understanding of both what a dataset contains and what domain
it belongs to.
This representation is the foundation on which vertical AI is built: a DLM
fine-tunes to any industry domain, composes into agents and tools, and operates
directly on raw data, with no preprocessing pipeline, no serialization, and no
engineering overhead between the raw data and the model.
It first determines what data it is analyzing, then predicts from it.
It is the native tabular understanding layer: the same role a language model plays
for text and a vision model plays for images.

We present \textbf{Schema-1}, a 140M parameter DLM trained on more than 2.3M
synthetic and real-world tabular datasets.

The contributions of this work are as follows.
\begin{itemize}
  \item We define the Data Language Model as a formal category of foundation model,
    with necessary and sufficient conditions that distinguish it from language models
    and tabular foundation models (Section~\ref{sec:dlm}).
  \item We present Schema-1, a 140M parameter DLM, and characterize its input
    structure, outputs, and operating properties (Section~\ref{sec:schema1}).
  \item We introduce \textbf{blind dataset sector classification} as a benchmark
    task specific to DLMs: given raw cell values with no metadata, identify the
    industry sector of the dataset, vertical-agnostically, across a broad range
    of human economic and scientific activity.
    No prior tabular model participates in this task.
    Schema-1 achieves 91.4\% top-1 accuracy.
  \item We demonstrate that Schema-1 ranks first on every benchmark evaluated,
    outperforming tree-based methods, AutoML ensembles, tabular foundation models,
    large language models, and classical statistical baselines, and establish through
    the column-agnostic evaluation that the imputation advantage is attributable to
    structural co-distributional learning rather than semantic signals in column
    identifiers, distinguishing data structure understanding from world knowledge
    derived from text (Section~\ref{sec:evaluation}).
\end{itemize}

\section{The Status Quo}
\label{sec:status_quo}

Before DLMs, three research directions addressed the use of tabular data in AI
systems.
Each represents a meaningful advance for predictive modeling and data science.
Tabular data is the primary information substrate on which enterprise decisions are
made, and all three directions serve that substrate well within the paradigm for
which they were designed.
The paradigm, now shifting toward vertical AI applications, purpose-built for
specific industries, and toward AI agents that must reason autonomously over
enterprise data across domains, surfaces a structural gap that all three directions
share.
In the classical path, raw data must be cleaned, normalized, encoded, and
feature-engineered into a numerical form before any model can train on it.
The model works on a processed proxy of the data, starts from scratch on every
new dataset, and carries no representation of what the data represents.
Any upstream schema change, a new column, a renamed field, a switched data source,
breaks the pipeline.
Tabular foundation models reduce the cold-start cost of the classical path but
preserve its fundamental output boundary: the output remains a probability vector
with no account of what the input dataset represents.
In the language model path, tables must be serialized into text before an LLM
can consume them: via text-to-SQL translation, row-by-row flattening, or
CSV and JSON wrapping.
Serialization destroys the joint distributional structure between columns, hits
context limits on any real-world dataset, and produces a model operating on
a text representation of data rather than data itself.
All three exist for the same reason: the AI at the end does not understand
tabular data natively.
All three impose compounding structural loss at every
stage~\cite{grinsztajn2022tree,shwartziv2022tabular}.

\paragraph{The hidden cost: domain specification.}
Before any of these three directions can be applied, a step must be completed
that no foundation-model approach has been designed to internalize: someone must
determine what the data actually is.
In practice this means an engineer or data scientist examines the column
distributions, resolves ambiguous or undocumented field names (internal codes like
\texttt{val\_B} or \texttt{metric\_14} are routine in production systems), consults
subject-matter experts to understand what values are meaningful, and translates that
understanding into pipeline decisions: which column is the target, which features
are relevant, which preprocessing steps are appropriate for this domain.
This data understanding step precedes every model training run on every new data
source and must be repeated in full when the source changes.
At the scale of a large enterprise operating across dozens of data sources and
business units, this manual process is a significant recurring cost.
For autonomous AI systems that encounter data sources not known at design time, it
is not a cost that can be managed: it is a dependency on a human that the system
cannot proceed without.
No prior tabular modeling approach was designed to replace this step.
The DLM paradigm places it inside the model.

\begin{figure}[H]
  \centering
  \begin{tikzpicture}[
    srcbox/.style={
      rectangle, draw=black!45, fill=gray!6,
      text width=1.6cm, align=center, minimum height=0.58cm,
      font=\scriptsize, inner sep=4pt, rounded corners=2pt
    },
    humanbox/.style={
      rectangle, draw=black!60, fill=gray!12,
      text width=1.55cm, align=center, minimum height=0.72cm,
      font=\scriptsize\itshape, inner sep=4pt, rounded corners=2pt
    },
    pipebox/.style={
      rectangle, draw=black!40, fill=gray!5,
      text width=2.0cm, align=center, minimum height=0.90cm,
      font=\scriptsize, inner sep=5pt, rounded corners=2pt
    },
    modelbox/.style={
      rectangle, draw=black!60, fill=white,
      text width=1.55cm, align=center, minimum height=0.58cm,
      font=\scriptsize\bfseries, inner sep=4pt, rounded corners=2pt
    },
    outbox/.style={
      rectangle, rounded corners=4pt, draw=black!50, fill=white,
      text width=1.55cm, align=center, minimum height=0.58cm,
      font=\scriptsize, inner sep=4pt
    },
    arr/.style={->, >=Stealth, line width=1.0pt, black!55},
    garr/.style={->, >=Stealth, line width=0.65pt, black!22},
  ]

  \node[srcbox]   (s1) at (0,    0) {Raw data\\source~1};
  \node[humanbox] (h1) at (2.1,  0) {domain\\specification\\(human)};
  \node[pipebox]  (p1) at (4.35, 0) {clean \textbar\ impute\\normalize \textbar\ encode\\feature eng.};
  \node[modelbox] (m1) at (6.6,  0) {train\\model};
  \node[outbox]   (o1) at (8.55, 0) {prediction};

  \draw[arr] (s1.east) -- (h1.west);
  \draw[arr] (h1.east) -- (p1.west);
  \draw[arr] (p1.east) -- (m1.west);
  \draw[arr] (m1.east) -- (o1.west);

  \node[srcbox, draw=black!18, fill=gray!3, text=black!28]
        (s2) at (0,    -1.55) {Raw data\\source~2};
  \node[humanbox, draw=black!18, fill=gray!5, text=black!25]
        (h2) at (2.1,  -1.55) {domain\\specification\\(human)};
  \node[pipebox, draw=black!15, fill=gray!3, text=black!22]
        (p2) at (4.35, -1.55) {clean \textbar\ impute\\normalize \textbar\ encode\\feature eng.};
  \node[modelbox, draw=black!18, fill=white, text=black!22]
        (m2) at (6.6,  -1.55) {train\\model};
  \node[outbox, draw=black!18, text=black!22]
        (o2) at (8.55, -1.55) {prediction};

  \draw[garr] (s2.east) -- (h2.west);
  \draw[garr] (h2.east) -- (p2.west);
  \draw[garr] (p2.east) -- (m2.west);
  \draw[garr] (m2.east) -- (o2.west);

  \node[draw=black!28, dashed, rounded corners=4pt,
        fit=(h1)(m2), inner sep=5pt] (pipebox) {};
  \node[font=\scriptsize\bfseries, text=black!42,
        above=3pt of pipebox] {bespoke pipeline, rebuilt from scratch per source};

  \node[font=\tiny, text=black!42] at (4.35, -2.55)
    {every new data source requires a complete pipeline rebuilt from scratch};

  \end{tikzpicture}
  \caption{
    The standard tabular ML pipeline.
    Domain specification is a manual human step: before any model can be trained,
    an engineer must determine what the data represents, resolve ambiguous field
    names, select the target column, and configure preprocessing accordingly.
    Every new data source requires rebuilding the full pipeline from scratch.
    Compare with Figure~\ref{fig:schema1_stack}.
  }
  \label{fig:status_quo_pipeline}
\end{figure}
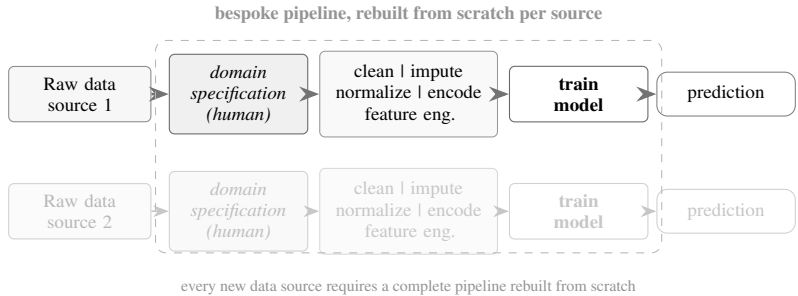

\paragraph{Classical methods and AutoML.}
Gradient-boosted decision trees, including XGBoost~\cite{chen2016xgboost},
LightGBM~\cite{ke2017lightgbm}, and CatBoost~\cite{prokhorenkova2018catboost},
remain strong baselines on structured prediction benchmarks.
AutoML systems such as AutoGluon~\cite{erickson2020autogluon} extend these with
automated ensembling and model selection.
These methods require preprocessing pipelines (normalization, encoding, imputation)
constructed and maintained externally to the model.
They learn conditional distributions within a given table and acquire no knowledge
that transfers to understanding the domain of a new one.
Every new dataset is a cold start.

\paragraph{Tabular foundation models.}
TabPFN~\cite{hollmann2023tabpfn} introduced prior-data fitted networks for tabular
classification, demonstrating that a transformer pretrained on synthetic data
generalizes to unseen classification tasks without task-specific training, a
significant advance over the cold-start regime.
Subsequent models including TabPFN-2.5~\cite{grinsztajn2025tabpfn25},
TabICLv2~\cite{qu2026tabicl}, and ConTextTab~\cite{spinaci2025contexttab}
extended this to larger datasets, regression, and richer input encodings.
These models eliminate the cold start but preserve the same output boundary:
a probability vector, with no account of what the input dataset represents.
Column names, when encoded, are consumed as input tokens without any corresponding
signal in the output.
The preprocessing requirement is reduced but not eliminated; the modality gap
remains fully intact.

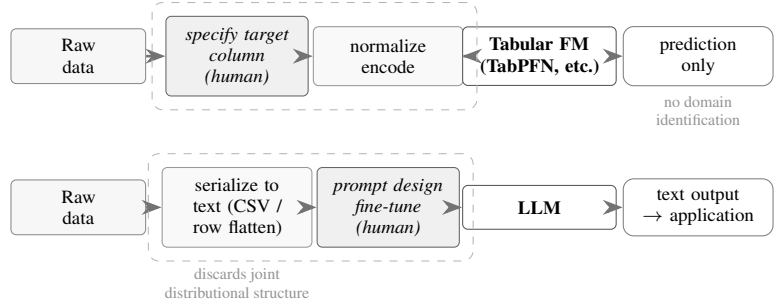
\begin{figure}[H]
  \centering
  \begin{tikzpicture}[
    srcbox/.style={
      rectangle, draw=black!45, fill=gray!6,
      text width=1.5cm, align=center, minimum height=0.58cm,
      font=\scriptsize, inner sep=4pt, rounded corners=2pt
    },
    humanbox/.style={
      rectangle, draw=black!60, fill=gray!12,
      text width=1.6cm, align=center, minimum height=0.78cm,
      font=\scriptsize\itshape, inner sep=4pt, rounded corners=2pt
    },
    pipebox/.style={
      rectangle, draw=black!40, fill=gray!5,
      text width=1.7cm, align=center, minimum height=0.78cm,
      font=\scriptsize, inner sep=4pt, rounded corners=2pt
    },
    modelbox/.style={
      rectangle, draw=black!65, fill=white,
      text width=1.75cm, align=center, minimum height=0.58cm,
      font=\scriptsize\bfseries, inner sep=4pt, rounded corners=2pt
    },
    outbox/.style={
      rectangle, rounded corners=4pt, draw=black!50, fill=white,
      text width=1.7cm, align=center, minimum height=0.58cm,
      font=\scriptsize, inner sep=4pt
    },
    arr/.style={->, >=Stealth, line width=1.0pt, black!55},
  ]

  \node[srcbox]   (t_s) at (0.0, 0) {Raw\\data};
  \node[humanbox] (t_h) at (2.1, 0) {specify target\\column\\(human)};
  \node[pipebox]  (t_p) at (4.1, 0) {normalize\\encode};
  \node[modelbox] (t_m) at (6.1, 0) {Tabular FM\\(TabPFN, etc.)};
  \node[outbox]   (t_o) at (8.2, 0) {prediction\\only};

  \draw[arr] (t_s.east) -- (t_h.west);
  \draw[arr] (t_h.east) -- (t_p.west);
  \draw[arr] (t_p.east) -- (t_m.west);
  \draw[arr] (t_m.east) -- (t_o.west);

  \node[draw=black!28, dashed, rounded corners=4pt,
        fit=(t_h)(t_p), inner sep=5pt] (tbox) {};

  \node[font=\tiny, text=black!42, align=center] at (8.2, -0.72)
    {no domain\\identification};

  \node[srcbox]    (l_s)   at (0.0, -2.0) {Raw\\data};
  \node[pipebox]   (l_ser) at (2.1, -2.0) {serialize to\\text (CSV /\\row flatten)};
  \node[humanbox]  (l_pe)  at (4.1, -2.0) {prompt design\\fine-tune\\(human)};
  \node[modelbox]  (l_m)   at (6.1, -2.0) {LLM};
  \node[outbox]    (l_o)   at (8.2, -2.0) {text output\\$\to$ application};

  \draw[arr] (l_s.east)    -- (l_ser.west);
  \draw[arr] (l_ser.east)  -- (l_pe.west);
  \draw[arr] (l_pe.east)   -- (l_m.west);
  \draw[arr] (l_m.east)    -- (l_o.west);

  \node[draw=black!28, dashed, rounded corners=4pt,
        fit=(l_ser)(l_pe), inner sep=5pt] (lbox) {};

  \node[font=\tiny, text=black!42, align=center] at (2.1, -3.0)
    {discards joint\\distributional structure};

  \end{tikzpicture}
  \caption{
    Builder pipelines for tabular foundation models and LLM-based approaches.
    Tabular foundation models reduce training cold-start but still require target
    column specification and a preprocessing step; their output is a prediction
    with no domain identification.
    LLM-based approaches replace preprocessing with serialization, converting
    tables to text sequences before the model operates; serialization discards
    the joint distributional structure between columns and ties the model to
    human-readable metadata.
    The dashed boxes mark the required steps each approach cannot bypass.
    Schema-1 eliminates both.
  }
  \label{fig:prior_pipelines}
\end{figure}

\paragraph{Large language models on tabular data.}
TabLLM~\cite{hegselmann2023tabllm}, TableGPT2~\cite{su2024tablegpt},
TabuLa-8B~\cite{gardner2024tabula}, and related approaches apply large language
models to tabular prediction by serializing tables into natural language.
These systems bring semantic reasoning to the problem and perform competitively
where column semantics carry predictive signal.
Serialization introduces a structural cost: converting a table to a token sequence
discards the joint distributional relationships between columns that are, in many
prediction tasks, the primary signal.
Performance degrades substantially when column identifiers are removed or replaced
with non-semantic labels~\cite{gardner2024tabula,spinaci2025contexttab}, confirming
that these systems depend on human-readable metadata that is frequently absent or
unreliable in production settings.
Figure~\ref{fig:prior_pipelines} contrasts the tabular-foundation-model and
LLM-based pipelines.

\paragraph{The shared limitation.}
The gap shared across all three paths is not a missing feature.
It is a missing paradigm.
No existing model class was designed with the requirement that understanding the
domain of a dataset is a first-class training objective.
In classical methods, the domain is specified by the engineer who built the
pipeline.
In tabular foundation models, the domain is irrelevant to the training objective.
In LLM-based approaches, the domain is inferred from column names in the prompt,
not from the data itself.
No prior approach treats domain identification as a primary model output: in each
case it is handled upstream, by the human assembling the pipeline.
For predictive modeling and data science workflows, this arrangement is workable:
a human is present to supply the domain context and assemble the pipeline.
For vertical AI applications that must operate on raw enterprise data across
multiple domains without human-in-the-loop pipeline assembly, and for AI agents
that must autonomously navigate datasets from industries the system designer did
not anticipate, treating domain identification as an upstream human responsibility
is not a solvable engineering problem.
It is a category error: the system requires a capability the model class was never
designed to provide.
A prominent ICML 2024 position formalizes this gap and calls for the development
of foundation models that can contextualize datasets rather than only predict
from them~\cite{vanbreugel2024tabular}.
The DLM paradigm we define in Section~\ref{sec:dlm} addresses a closely related
but more fundamental capability: not relating a dataset to a corpus of prior
datasets, but identifying the informational content of a dataset as a primary
inference output from its own structural properties, without external labels,
metadata, or a hand-assembled pipeline.

\section{Data Language Models}
\label{sec:dlm}

We define the Data Language Model as a new class of foundation model.
The definition emerges from a practical requirement: vertical AI applications
and AI agents that operate on enterprise tabular data need a layer that
operates directly on raw tables, without a bespoke preprocessing pipeline for
each new domain or data source.
The definition is stated in terms of three conditions.
Each condition excludes at least one prior paradigm; together they define a
capability that none of the prior model classes considered here were designed to
provide.

\begin{definition}[Data Language Model]
\label{def:dlm}
A \textbf{Data Language Model} is a foundation model that simultaneously
satisfies the following three conditions:
\begin{enumerate}
  \item \textbf{Multi-signal native ingestion.}
    The model processes raw tabular inputs directly, encoding column semantics,
    per-column distributional properties, cell values, and missing value structure
    within a single unified representation where each signal informs the
    interpretation of the others.
    Valid outputs are produced when column semantic signals are absent.

  \item \textbf{Dataset-level contextual inference.}
    The model produces, as a primary inference result, a structured identification
    of the domain or sector that generated the input dataset, derived entirely from
    the data's own structural and distributional properties, without external labels,
    metadata, or world knowledge.
    This identification is a co-equal output of the same inference step that produces
    numerical predictions, not a post-hoc annotation.

  \item \textbf{Metadata-independent operation.}
    Removing all column names and dataset metadata causes no more than a small,
    quantified degradation in predictive performance.
    Column semantic signals are one input pathway among several, not a necessary
    condition for inference.
\end{enumerate}
\end{definition}

The three conditions describe the full input-output contract: Condition~1 defines
what the model accepts, Condition~2 defines what it returns, and Condition~3
specifies the robustness guarantee under degraded input.

\paragraph{Condition elaborations.}
Condition~1 requires that the four input signals it names, column semantics,
per-column distributional properties, cell values, and missing value structure,
be jointly processed so that each informs the interpretation of the others
throughout the model's computation.

Condition~2 extends to multi-source inputs.
When multiple datasets are submitted simultaneously, the model identifies each
independently from its own distributional structure and surfaces shared distributional
signatures, correlated column distributions, and structural correspondences across
sources as additional outputs.
Each dataset may originate from a different enterprise system, carry a different
schema, use different column names, and arrive in a different encoding convention:
no cross-dataset labels, join keys, relational metadata, lookup tables, or
normalization across sources is required or consulted at any stage.
This is a capability with no equivalent among the prior tabular model classes
considered here, all of
which operate on a single labeled dataset with a specified target column and have
no defined behavior on multi-dataset, multi-source, label-free input.

Condition~3 is quantified formally.
Let $M_{\mathrm{full}}$ denote the model evaluated with all input pathways active
and $M_\emptyset$ the same model with the column semantic pathway zeroed.
Condition~3 requires
\begin{equation*}
  \mathrm{AUC}(M_{\mathrm{full}}) - \mathrm{AUC}(M_\emptyset)
    \;\leq\; \varepsilon
\end{equation*}
for a small, pre-specified $\varepsilon$.
In Schema-1, $\varepsilon = 0.0117$ (98.8\% retention of predictive
performance), established empirically in Section~\ref{sec:evaluation}.

Table~\ref{tab:dlm_conditions} shows which conditions each prior paradigm satisfies.
Among the paradigms surveyed, none satisfies all three simultaneously.

\begin{table}[H]
  \centering
  \resizebox{\linewidth}{!}{%
  \begin{tabular}{lccc}
    \toprule
    Paradigm & C1: Multi-signal & C2: Dataset-level & C3: Metadata- \\
             & native ingestion & contextual inference & independent \\
    \midrule
    Tree-based methods
      (XGBoost, LightGBM)          & $\times$ & $\times$ & \checkmark \\
    Numerical tabular foundation models
      (TabPFN)                     & $\times$ & $\times$ & \checkmark \\
    Semantics-aware tabular models
      (ConTextTab)                 & $\times$\textsuperscript{$\dagger$}    & $\times$ & $\times$ \\
    Serialization-based LLM approaches
      (TabLLM, TableGPT2, TabuLa-8B) & $\times$ & $\times$ & $\times$ \\
    \textbf{Schema-1 (DLM)}         & \checkmark & \checkmark & \checkmark \\
    \bottomrule
  \end{tabular}}
  \vspace{0.5ex}
  \begin{flushleft}
  \footnotesize
  \textsuperscript{$\dagger$}~Semantics-aware models do encode column names, but
  Condition~1 requires that all four input signals be processed jointly so that
  each informs the interpretation of the others. In ConTextTab the semantic
  and structural pathways are encoded independently, so the joint
  requirement is not met.
  C2 requires that domain identification be derived from the data's own
  structural and distributional properties, as a primary model output.
  Serialization-based LLM approaches produce text responses grounded in
  world knowledge from training corpora; they do not produce a domain
  identification derived from the input table's own distributional content,
  and require serialization rather than native ingestion.
  \end{flushleft}
  \caption{
    Satisfaction of the three DLM conditions across model paradigms.
    A checkmark indicates the condition is satisfied; a cross indicates it is not.
    Schema-1 is the only system satisfying all three.
  }
  \label{tab:dlm_conditions}
\end{table}

\paragraph{The training objective distinction.}
Every prior tabular model class is optimized to answer one question from data:
given these features, what is the target value?
A DLM is additionally required to answer a second question from the same input:
given these cells, what domain generated this data?
These are structurally different questions, and requiring both to be answerable
simultaneously from a single inference pass produces a qualitatively
different kind of model.
The first question is answered by learning conditional distributions within a
dataset.
The second is answered by learning what distributional signatures characterize
each domain across datasets.
A model that has learned both encodes not only within-dataset structure but
cross-dataset identity: it represents where in the space of all possible
data-generating processes the current table lives.
No prior tabular model was trained with the second requirement, and none
produce the second answer.
This is not a capability gap in existing models: it is outside the scope of the
prediction task they were designed for.
It is a new training objective, and the new training objective is what defines
the new model class.

\paragraph{On the semantics of ``understanding.''}
A Data Language Model comprehends any raw table on the spot.
It requires no prior knowledge of the vertical, no preprocessing, no target
column specification, and no tolerance for clean or complete input.
A DLM receives a raw dataset, whatever its domain, complexity, noise level, or
pattern of missing values, and produces domain identification and predictions in
a single inference step.
This is vertical agnosticism as a first-class model property, not as a
post-training specialization or configuration option.
The same model weights that operate on a critical care ICU record also operate
on a semiconductor yield log or a cross-border payment stream, with no
reconfiguration between them.

The property that makes this possible is structural comprehension at the
dataset level.
A DLM does not match input tables against memorized domain templates.
It reads the distributional geometry of the data itself: value range
relationships, covariance patterns, missingness structure, and the joint
distributional profile across all columns simultaneously.
From this alone, and in the presence of incompleteness, noise, and naming
ambiguity, the model places the dataset correctly in the space of all
domains of human data-producing activity.
No world knowledge is consulted.
No external labels or metadata are referenced.

A skilled analyst demonstrates the same capacity.
Given an unfamiliar dataset with no documentation, the analyst reads its column
distributions, value scales, and correlation structure, and within moments
places it in a domain context, before any modeling, without being told
what the data represents.
That is what we recognize as genuine data intelligence in a practitioner.
A DLM performs the same inferential operation, from the same evidential basis,
across the full scope of human knowledge-producing activity simultaneously,
without human intervention to supply the context the model would otherwise lack.
The understanding is in the model.
That is the intelligence the DLM paradigm introduces into the AI stack.

\paragraph{What the DLM paradigm enables for builders.}
The practical consequence of the DLM conditions is that a DLM can function as
the foundation model on which vertical AI applications and autonomous agents
are built, without preprocessing, without domain configuration, and without
a human-in-the-loop to supply context the model cannot derive itself.
An AI agent equipped with a DLM can receive multiple raw datasets from
different enterprise sources simultaneously, determine what domain each
represents, automatically surface the correlational relationships between them,
produce predictions, and pass all outputs to downstream reasoning
in a single inference step with no pipeline assembly required.
A vertical AI application can be built against a DLM and deployed across new
enterprise verticals without re-training, re-labeling, or reconfiguration for
each new data source encountered at runtime.
None of these workflows are achievable with any prior tabular model class.
Tree-based methods and tabular foundation models require a specified target
column and labeled training data; they have no defined behavior on unlabeled,
multi-dataset, context-free input.
LLM-based approaches require serialization that destroys joint distributional
structure and produce responses grounded in world knowledge rather than in the
structural content of the data at hand.
The DLM paradigm closes the gap: raw tables enter; domain identification,
predictions, and cross-dataset correlations exit.

\paragraph{Relationship to language models and vision models.}
Each major data modality has produced a class of foundation models that
eliminated the preprocessing pipeline previously required to operate on it.
For language, large language models replaced the assembly of tokenization,
embedding computation, named entity recognition, and manual feature extraction:
text enters the model, understanding exits.
For images, diffusion models and vision encoders replaced hand-engineered
feature pipelines: raw pixels enter, generation or representation exits.
For audio, foundation models replaced spectral feature engineering and
task-specific signal processing chains.
In every case, the model understands the modality natively, and that native
understanding is what makes the pipeline unnecessary.
Tabular data has not had this.
Every prior approach requires a preprocessing pipeline before inference can
begin: normalization, encoding, imputation, feature engineering.
That pipeline exists because no model has understood raw tabular structure
natively, in the same way text and images are understood natively by the
foundation models built for them.
Data Language Models are that class for structured data.
The column named \texttt{albumin\_g\_per\_L} and the column named
\texttt{feature\_03} are computationally indistinguishable to a model whose
only objective is to predict a target column.
A DLM resolves this anomaly by introducing the second training objective:
identify what this dataset is, not only what to predict from it.
Domain identification and prediction exit the model as co-equal inference
results produced in a single inference step.
A DLM is a model class defined by the requirement that understanding the content
of the input is a primary training objective, co-equal with prediction.
Text and images each acquired this layer decades ago, and the capability
unlocked by each has been foundational to everything built on top of it since.
Tabular data is acquiring it now.

\section{Schema-1}
\label{sec:schema1}

Schema-1 is the first concrete instantiation of the Data Language Model paradigm
defined in Section~\ref{sec:dlm}.
It is the foundation model for vertical AI and agentic AI development: the layer
on which AI models, agents, and vertical applications can be built directly on raw
tabular data, with native modality understanding for the structured-data substrate.
Row-level prediction, the task on which tabular benchmarks evaluate, is one
application of this layer, not its definition.
Its design follows a single organizing principle: the raw table enters the model
directly, and the model determines what the data is and what can be predicted from
it, with no preprocessing pipeline, no feature engineering, and no external labels
required before inference.
Schema-1 ingests four input signals: column semantics, per-column
distributional summaries, raw cell values, and missing value structure, fused into
a unified representation so that each signal informs the interpretation of the
others.
When multiple datasets from different enterprise sources are presented
simultaneously, each passes through the same four input pathways
independently: no shared schema, normalized column space, foreign key linkage,
relational metadata, or alignment across sources is required or assumed.
Column names are treated as an optional enhancement rather than a requirement:
when absent, the model operates on the remaining three signals with a measured
and bounded degradation.
Missing values are encoded as structural signals, not errors to be removed before
the model can run.
Both outputs, the domain identification and the numerical prediction, are produced
in a single inference step, making dataset understanding and prediction
a single operation rather than a pipeline.
Schema-1 also supports sequential fine-tuning: it accumulates task knowledge
without discarding prior tasks, making it viable as a persistent tabular
understanding layer that grows with use.
The training data, modality structure, and deployment configuration are described
in the subsections below.

\paragraph{Modality structure.}
Schema-1 implements Condition~1 of the DLM definition through four input signals,
each of which positions every cell as a point in a learned vector space.
The representation of each cell is shaped not only by its own four input signals
but by its relationships to all other cells across the full structure of the dataset.
The model therefore captures joint distributional structure from the data itself,
without any external schema or metadata.

\begin{itemize}
  \item \textbf{Column semantics.}
    A grounded encoding of column identifiers and their semantic content.
    When column names are absent, the corresponding input is zeroed and the model
    operates on the remaining three signals.
    Empirically, removing column names causes a drop of 0.0117 in mean ROC-AUC (from 0.9435
    to 0.9318), 98.8\% retention of predictive performance
    (Section~\ref{sec:evaluation}).

  \item \textbf{Per-column distributional summaries.}
    Summary statistics computed per column directly from cell values, without any
    user-provided metadata: central tendency, dispersion, range, the ratio of
    numeric to total values, unique value density, and the number of observed
    values.
    This signal is unaffected by missingness in individual cells and does not
    require that column names be present or semantically meaningful.

  \item \textbf{Cell values.}
    Raw numeric and categorical cell values across all columns.
    Numeric and categorical values are each represented within a unified cell
    encoding before the model's primary reasoning stages.

  \item \textbf{Missing value structure.}
    A structural encoding of the pattern of present and absent values across rows
    and columns.
    Schema-1 treats missing values as meaningful structural signals rather than
    errors to be preprocessed away.
    No imputation step is required before inference.
\end{itemize}

Each signal carries independent predictive information: no single pathway is a
necessary condition for inference.

\paragraph{Multi-source ingestion.}
When Schema-1 receives datasets from multiple enterprise sources in a single
inference call, the same four-pathway structure applies to each independently.
No schema alignment, shared column namespace, join keys, or external metadata
relating one source to another is required: cross-dataset distributional
relationships are derived from the data itself.

\paragraph{Training data.}
Schema-1 is trained on 2,307,000 datasets in total: 2,000,000 synthetic datasets
generated from a controlled sector-specific schema covering a broad range of
human economic and scientific sectors, and 307,000 real-world datasets drawn from
public and domain-specific sources.
The two components serve distinct purposes.
The synthetic component provides coverage breadth: structurally grounded datasets
per sector ensure that the model has seen representative distributional
profiles, cell value geometries, and column covariance patterns across the
complete sector range, including sectors for which large real-world labeled
datasets do not exist.
The real-world component anchors the model to naturalistic data collection
artifacts: irregular distributions, measurement noise, encoding conventions, and
missing value patterns that synthetic generation does not fully replicate.
Together, the two components provide both the breadth required for cross-domain
sector identification and the distributional realism required for transfer to
production data.
The sector taxonomy comprises 10,000 industry sectors spanning a broad range of
human economic and scientific activity: from primary-industry sub-sectors (mussel
farming, ruby mining) to advanced technology domains (augmented reality,
semiconductor design).
All datasets used in the Section~\ref{sec:evaluation} benchmarks were excluded
from the training corpus by identifier-level blocklisting prior to training, to
prevent train-test overlap.

\paragraph{Training objective.}
Schema-1 is trained to jointly satisfy all three conditions of
Definition~\ref{def:dlm}.
The training objective is designed so that satisfying each condition reinforces
the others: a model that correctly identifies domains must have learned the
distributional structure that also improves prediction and missing value handling.

\paragraph{Fine-tuning.}
Fine-tuning adapts Schema-1 to a specific enterprise task on customer-supplied
labeled data.
The base model weights remain frozen throughout; each fine-tuning run produces a
customer-specific isolated checkpoint that is the sole model used in that
deployment.
No fine-tuning job modifies the shared base model, and no customer's data or
checkpoint is accessible from any other customer's deployment.
All evaluations in Section~\ref{sec:evaluation} use Schema-1 in this
configuration.

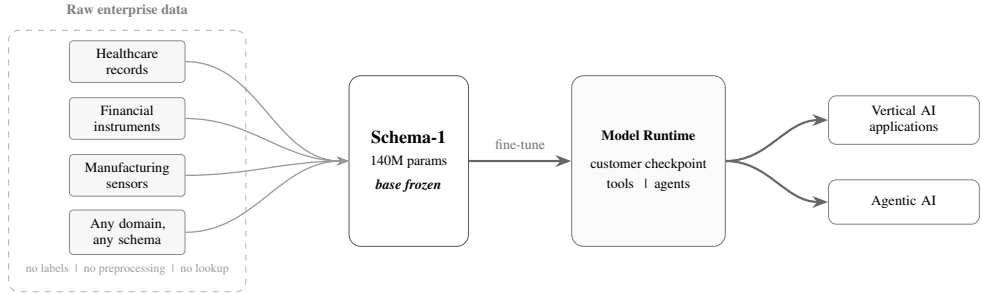
\begin{figure}[H]
  \centering
  \resizebox{0.92\textwidth}{!}{%
  \begin{tikzpicture}[
    databox/.style={
      rectangle, draw=black!45, fill=gray!6,
      text width=1.75cm, align=center, minimum height=0.60cm,
      font=\scriptsize, inner sep=4pt, rounded corners=2pt
    },
    mainbox/.style={
      rectangle, rounded corners=6pt, draw=black!65, fill=white,
      align=center, font=\scriptsize\bfseries, inner sep=6pt
    },
    runtimebox/.style={
      rectangle, rounded corners=6pt, draw=black!55, fill=gray!4,
      align=center, font=\scriptsize, inner sep=6pt,
      minimum width=2.7cm
    },
    appbox/.style={
      rectangle, rounded corners=4pt, draw=black!50, fill=white,
      text width=2.3cm, align=center, minimum height=0.78cm,
      font=\scriptsize, inner sep=5pt
    },
    arr/.style={->, >=Stealth, line width=1.1pt, black!60},
    thin arr/.style={->, >=Stealth, line width=0.65pt, black!40},
    lbl/.style={font=\scriptsize, text=black!55, midway, above=2pt},
  ]

  \node[databox] (d1) at (0,  1.5) {Healthcare\\records};
  \node[databox] (d2) at (0,  0.5) {Financial\\instruments};
  \node[databox] (d3) at (0, -0.5) {Manufacturing\\sensors};
  \node[databox] (d4) at (0, -1.5) {Any domain,\\any schema};

  \node[font=\tiny, text=black!38] (tag) at (0, -2.15)
    {no labels \;|\; no preprocessing \;|\; no lookup};

  \node[draw=black!30, dashed, rounded corners=5pt,
        fit=(d1)(d4)(tag), inner sep=5pt] (datasrc) {};
  \node[font=\scriptsize\bfseries, text=black!50,
        above=3pt of datasrc] {Raw enterprise data};

  \node[mainbox, minimum height=3.0cm, minimum width=2.1cm,
        right=1.8cm of datasrc] (s1)
    {{\small\bfseries Schema-1}\\[3pt]\normalfont 140M params\\[4pt]
     \itshape base frozen};

  \foreach \d in {d1,d2,d3,d4}{
    \draw[thin arr] (\d.east) to[out=0,in=180] (s1.west);
  }

  \node[runtimebox, minimum height=3.0cm,
        right=1.8cm of s1] (rt)
    {\bfseries Model Runtime\\[6pt]
     \normalfont customer checkpoint\\[2pt]
     tools \enspace|\enspace agents};

  \draw[arr] (s1.east) --
    node[font=\scriptsize, text=black!55, midway, above=2pt] {fine-tune}
    (rt.west);

  \node[appbox, right=1.8cm of rt, yshift= 0.72cm] (vapp)
    {Vertical AI\\applications};
  \node[appbox, right=1.8cm of rt, yshift=-0.72cm] (agent)
    {Agentic AI};

  \draw[arr] (rt.east) to[out=0,in=180] (vapp.west);
  \draw[arr] (rt.east) to[out=0,in=180] (agent.west);

  \end{tikzpicture}}%
  \caption{
    Schema-1 as the foundation model for vertical and agentic AI.
    Multiple raw enterprise datasets from any domain enter Schema-1 directly,
    without labels, preprocessing, or external lookup.
    Fine-tuning on customer-supplied labeled data produces an isolated
    customer-specific checkpoint; the base model is never modified and no
    customer data is accessible from any other deployment.
    The checkpoint runs inside a Model Runtime enabling multi-tool and agentic
    development.
    Compare with Figure~\ref{fig:status_quo_pipeline}.
  }
  \label{fig:schema1_stack}
\end{figure}

\paragraph{Schema-1 as a foundation layer.}
Figure~\ref{fig:schema1_stack} shows how Schema-1 is positioned in a production AI
stack.
Each fine-tune job produces a customer-specific isolated checkpoint while the
base model remains frozen and unmodified, so every deployment is unique,
private, and independent.
The resulting model is served inside the Model Runtime, which exposes tool
interfaces and agent scaffolding: the layer on which vertical AI applications
and autonomous agents are built.

\section{Evaluation}
\label{sec:evaluation}

We evaluate Schema-1 across six benchmarks: OpenML-CC18 row-level prediction,
missing data robustness, tabular imputation quality, column-agnostic prediction
robustness, blind dataset sector classification, and sequential fine-tuning
retention.
Table~\ref{tab:summary} summarizes all results; sector classification is reported
at two precision levels and contributes two rows.
Schema-1 ranks first on every benchmark and on the large majority of
evaluation conditions reported.

All evaluations use Schema-1 in fine-tuning mode with the backbone frozen, with
the exception of the blind dataset sector classification benchmark, which uses
Schema-1 in sector classification mode.

\begin{table}[H]
  \centering
  \small
  \setlength{\tabcolsep}{5pt}
  \begin{tabular}{lccc}
    \toprule
    Benchmark & Schema-1 & Best competitor & Margin \\
    \midrule
    CC18 mean ROC-AUC (18 datasets)    & \textbf{0.9849} & 0.9339 (TabPFN+AG)  & $+$0.0510 \\
    Missing data mean ROC-AUC (0--70\%)& \textbf{0.9196} & 0.8933 (MIRRAMS)    & $+$0.0263 \\
    Imputation NRMSE ($\downarrow$)    & \textbf{0.163}  & 0.235 (Gemini~3.0)  & $-$31\%   \\
    Column-agnostic ROC-AUC (no names) & \textbf{0.9318} & 0.8658 (TabuLa-8B)  & $+$0.0660 \\
    Sector classification top-1        & \textbf{91.4\%} & 0.01\% (random)     & $+$91.4 pp    \\
    Sector classification top-5        & \textbf{97.0\%} & 0.05\% (random)     & $+$97.0 pp    \\
    Seq. fine-tuning retention         & \textbf{97.8\%} & 0\% (GBDTs: retrain)& $+$97.8 pp$^\dagger$ \\
    \bottomrule
  \end{tabular}
  \caption{
    Summary of all benchmark results.
    Schema-1 ranks first on every benchmark and on the large majority of
    evaluation conditions.
    $\dagger$: GBDTs (gradient-boosted decision trees, e.g., XGBoost, LightGBM,
    CatBoost) require full retraining per task; the margin is the absolute
    difference in retention, not a ratio.
  }
  \label{tab:summary}
\end{table}

\paragraph{OpenML-CC18 row-level prediction.}
CC18 is a widely used reference benchmark for tabular methods, with a stable
evaluation protocol that has been in use since 2022.
Schema-1 scores highest on all 18 datasets, setting a new performance ceiling
across the full CC18 suite.
On datasets representing real-world hard problems, predicting patient outcomes,
blood donation behavior, and software reliability, Schema-1 moves from roughly
correct to nearly perfect.

\textit{Methodology.}
We follow the original benchmark protocol exactly~\cite{hollmann2023tabpfn}: 18
numerical classification datasets from the OpenML-CC18 benchmarking
suite~\cite{bischl2021openml}, 10-fold stratified cross-validation, a
60-minute wall-clock training budget per fold, ROC-AUC one-vs-one
(prediction accuracy; 1.0 is perfect, higher is better) as the primary
metric, and standard per-fold normalization as specified in the original
protocol.
All external competitor per-dataset values are sourced from~\cite{hollmann2023tabpfn},
Table~2, using identical fold splits and dataset versions; means and ranks reported
in Table~\ref{tab:cc18} are computed over the 18-dataset subset on which Schema-1
is evaluated. No external competitor was re-run by SchemaLabs.
TabPFN-2.5~\cite{grinsztajn2025tabpfn25}, TabICLv2~\cite{qu2026tabicl}, and
subsequent variants have not published results under the Hollmann et al.\ protocol
and are not included.

\textit{Results.}
Schema-1 achieves mean ROC-AUC~0.9849, ranking first across all eight systems
on all 18 datasets.
The mean margin over the next-best model (TabPFN+AG, 0.9339) is 0.0510 ROC-AUC.
The largest absolute gains occur on datasets where all prior models exhibit
substantial error: \texttt{diabetes} (0.8247--0.8427 across all prior baselines
vs.\ Schema-1 at 0.9624), \texttt{blood-transfusion} (0.7144--0.7549
vs.\ 0.9428), \texttt{pc1} (0.8321--0.8761 vs.\ 0.9798), \texttt{pc3}
(0.8178--0.8373 vs.\ 0.9712), and \texttt{kc2} (0.8141--0.8346 vs.\ 0.9685).
Per-dataset results are in Table~\ref{tab:cc18}.

\begin{table}[H]
  \centering
  \footnotesize
  \setlength{\tabcolsep}{3.5pt}
  \resizebox{\textwidth}{!}{%
  \begin{tabular}{lccccccc|>{\columncolor{gray!9}}c}
    \toprule
    Dataset & LightGBM & CatBoost & XGBoost & ASKL2 & AutoGluon
            & TabPFN & TabPFN+AG & \textbf{Schema-1} \\
    \midrule
    balance-scale           & 0.9938 & 0.9245 & 0.9939 & 0.9970 & 0.9919
                            & 0.9973 & 0.9958 & \textbf{0.9996} \\
    mfeat-fourier           & 0.9786 & 0.9816 & 0.9803 & 0.9826 & 0.9843
                            & 0.9811 & 0.9838 & \textbf{0.9918} \\
    mfeat-karhunen          & 0.9979 & 0.9986 & 0.9983 & 0.9975 & 0.9987
                            & 0.9978 & 0.9985 & \textbf{0.9998} \\
    mfeat-morphological     & 0.9601 & 0.9629 & 0.9612 & 0.9671 & 0.9698
                            & 0.9669 & 0.9722 & \textbf{0.9872} \\
    mfeat-zernike           & 0.9716 & 0.9759 & 0.9735 & 0.9812 & 0.9908
                            & 0.9823 & 0.9901 & \textbf{0.9961} \\
    diabetes                & 0.8247 & 0.8383 & 0.8378 & 0.8343 & 0.8391
                            & 0.8410 & 0.8427 & \textbf{0.9624} \\
    vehicle                 & 0.9232 & 0.9302 & 0.9282 & 0.9504 & 0.9416
                            & 0.9589 & 0.9538 & \textbf{0.9842} \\
    analcatdata\_authorship & 0.9999 & 0.9999 & 0.9997 & 1.0000 & 0.9993
                            & 1.0000 & 1.0000 & \textbf{1.0000} \\
    pc4                     & 0.9301 & 0.9413 & 0.9291 & 0.9331 & 0.9428
                            & 0.9383 & 0.9440 & \textbf{0.9841} \\
    pc3                     & 0.8178 & 0.8247 & 0.8288 & 0.8265 & 0.8282
                            & 0.8373 & 0.8360 & \textbf{0.9712} \\
    kc2                     & 0.8141 & 0.8323 & 0.8227 & 0.8311 & 0.8242
                            & 0.8346 & 0.8321 & \textbf{0.9685} \\
    pc1                     & 0.8321 & 0.8600 & 0.8489 & 0.8527 & 0.8578
                            & 0.8761 & 0.8739 & \textbf{0.9798} \\
    wdbc                    & 0.9904 & 0.9931 & 0.9904 & 0.9947 & 0.9956
                            & 0.9964 & 0.9960 & \textbf{0.9994} \\
    qsar-biodeg             & 0.9126 & 0.9217 & 0.9191 & 0.9247 & 0.9276
                            & 0.9336 & 0.9336 & \textbf{0.9856} \\
    banknote-authentication & 1.0000 & 1.0000 & 1.0000 & 1.0000 & 1.0000
                            & 1.0000 & 1.0000 & \textbf{1.0000} \\
    blood-transfusion       & 0.7144 & 0.7403 & 0.7312 & 0.7504 & 0.7364
                            & 0.7549 & 0.7469 & \textbf{0.9428} \\
    steel-plates-fault      & 0.9626 & 0.9655 & 0.9656 & 0.9694 & 0.9666
                            & 0.9655 & 0.9687 & \textbf{0.9918} \\
    climate-model-sim.      & 0.9286 & 0.9344 & 0.9255 & 0.9291 & 0.9391
                            & 0.9415 & 0.9421 & \textbf{0.9832} \\
    \midrule
    \textbf{Mean ROC-AUC}   & 0.9196 & 0.9236 & 0.9241 & 0.9290 & 0.9297
                            & 0.9335 & 0.9339 & \textbf{0.9849} \\
    \textbf{Mean Rank}      & 7.06   & 5.06   & 6.22   & 4.50   & 4.17
                            & 3.17   & 2.72   & \textbf{1.00}   \\
    \textbf{Wins}           & 1/18   & 1/18   & 1/18   & 2/18   & 1/18
                            & 2/18   & 2/18   & \textbf{18/18}  \\
    \bottomrule
  \end{tabular}}
  \caption{
    OpenML-CC18 benchmark: ROC-AUC (one-vs-one) on 18 numerical classification
    datasets under 10-fold stratified cross-validation with a 60-minute
    wall-clock budget per fold, including standard per-fold normalization
    following the original protocol.
    LightGBM through TabPFN+AG: sourced from~\cite{hollmann2023tabpfn},
    Table~2, identical fold splits and dataset versions; no external model
    re-run by SchemaLabs.
    ASKL2: Auto-Sklearn~2.0~\cite{feurer2022autosklearn}.
    Mean Rank and Wins computed across all eight models in this table.
    \textbf{Bold}: best result per dataset.
  }
  \label{tab:cc18}
\end{table}

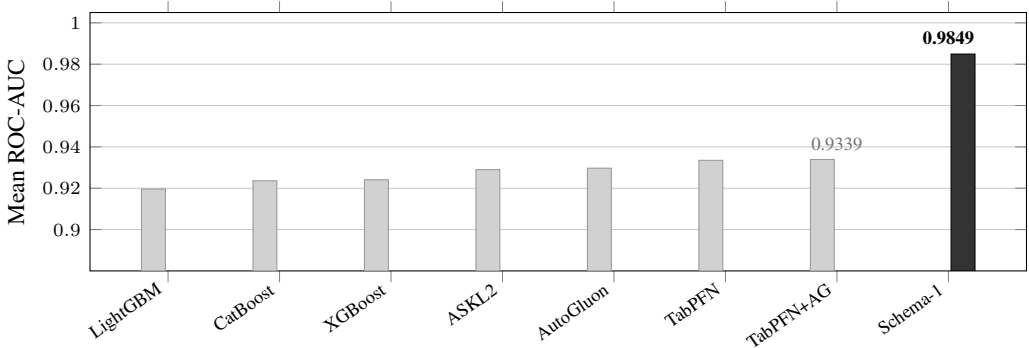
\begin{figure}[H]
  \centering
  \begin{tikzpicture}
    \begin{axis}[
      ybar,
      bar width=9pt,
      width=\columnwidth,
      height=5.0cm,
      ylabel={Mean ROC-AUC},
      ymin=0.88, ymax=1.005,
      ytick={0.9000,0.9200,0.9400,0.9600,0.9800,1.0000},
      yticklabel style={font=\scriptsize, /pgf/number format/.cd, fixed, precision=4},
      xtick={1,2,3,4,5,6,7,8},
      xticklabels={LightGBM,CatBoost,XGBoost,ASKL2,AutoGluon,
                   TabPFN,TabPFN+AG,Schema-1},
      xticklabel style={rotate=35,anchor=east,font=\scriptsize},
      ymajorgrids=true,
      grid style={line width=0.2pt,draw=gray!25},
      major grid style={line width=0.35pt,draw=gray!45},
      tick label style={font=\scriptsize},
      label style={font=\small},
    ]
    \addplot[fill=black!18,draw=black!40] coordinates {
      (1,0.9196)(2,0.9236)(3,0.9241)(4,0.9290)(5,0.9297)
      (6,0.9335)(7,0.9339)};
    \addplot[fill=black!80,draw=black] coordinates {(8,0.9849)};
    \node[above, font=\scriptsize, text=black!55] at (axis cs:7,0.9339) {0.9339};
    \node[above, font=\scriptsize\bfseries, text=black] at (axis cs:8,0.9849) {0.9849};
    \end{axis}
  \end{tikzpicture}
  \caption{
    Mean ROC-AUC on OpenML-CC18 (18 datasets, 10-fold CV).
    Schema-1 (rightmost, dark) at 0.9849.
    The gap between Schema-1 (0.9849) and second place (TabPFN+AG, 0.9339) is
    0.0510, larger than the entire range spanned by all prior competitors
    (0.9196--0.9339).
  }
  \label{fig:cc18_bar}
\end{figure}

\textit{What the numbers mean.}
CC18 has been the reference benchmark for tabular methods since 2022; the
systems most widely deployed in production (XGBoost, LightGBM, AutoGluon,
TabPFN) have been evaluated against it under a stable protocol.
Schema-1 ranked first on every single one of the 18 datasets.
The gap is not marginal: Schema-1's mean ROC-AUC is 0.9849, versus 0.9339 for
the previous best (TabPFN+AG).
On the five hardest real-world problems (patient outcomes, blood donation
behavior, software reliability), Schema-1 moves from a performance band of
0.71--0.88 to 0.94--0.98, not an incremental improvement within a prior
distribution but a distinct accuracy tier.
These results reflect a model that has learned the underlying structure of
tabular data in general, not just the specific datasets it was trained on,
achieved through the structural representation that cross-domain pretraining
on more than 2.3M datasets produces.
Figure~\ref{fig:cc18_bar} shows the mean ROC-AUC comparison across all models.

\paragraph{Missing data robustness.}
In production, data gaps are not edge cases; they are the norm.
Every real-world tabular pipeline must decide what to do when a fraction of
incoming values are absent: impute upstream, drop rows, retrain, or accept
degraded accuracy.
\textbf{Schema-1 at 70\% data missing outperforms every other system tested at
50\% missing.}
This eliminates an entire class of brittle data pipelines and the engineering cost
of maintaining them.

\textit{Methodology.}
Mean ROC-AUC is reported on 15 CC18 datasets under MCAR missingness (missing
values occur randomly, with no dependency on any feature in the dataset),
injected into the test set at six rates (0, 10, 20, 30, 50, and 70\%).
MAR (a dependency between existing features determines which values go
missing) and MNAR (missing values depend on themselves or unobserved
external data; the hardest case) conditions are evaluated separately in the
imputation benchmark below.
Training data remains complete in all conditions.
Schema-1 handles missing values natively inside the model; no imputation
preprocessing is applied before inference.
MIRRAMS~\cite{lee2025mirrams} is a deep learning framework designed specifically
for robustness under missingness distribution shifts;
TabPFN-2.5~\cite{grinsztajn2025tabpfn25} and XGBoost with KNN and mean imputation
serve as additional baselines.

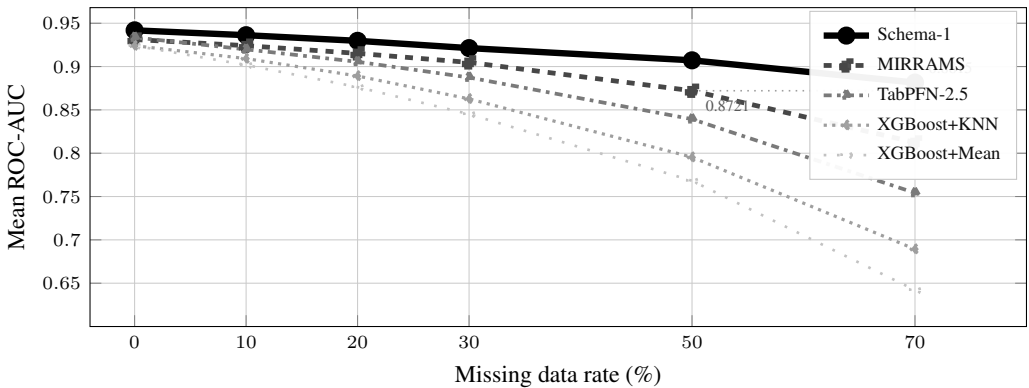
\begin{figure}[H]
  \centering
  \begin{tikzpicture}
    \begin{axis}[
      width=\columnwidth,
      height=5.8cm,
      xlabel={Missing data rate (\%)},
      ylabel={Mean ROC-AUC},
      xmin=-4, xmax=80,
      ymin=0.600, ymax=0.968,
      xtick={0,10,20,30,50,70},
      ytick={0.65,0.70,0.75,0.80,0.85,0.90,0.95},
      grid=both,
      grid style={line width=0.2pt, draw=gray!22},
      major grid style={line width=0.35pt, draw=gray!42},
      legend style={at={(0.99,0.99)}, anchor=north east, font=\scriptsize,
                    fill=white, fill opacity=0.95, draw=gray!45,
                    row sep=1pt, inner sep=4pt},
      legend cell align=left,
      tick label style={font=\scriptsize},
      label style={font=\small},
      clip=false,
    ]
    \addplot[black, line width=2.4pt, mark=*, mark size=2.2pt, solid]
      coordinates {(0,0.9418)(10,0.9362)(20,0.9297)(30,0.9213)(50,0.9072)(70,0.8815)};
    \addlegendentry{Schema-1}
    \addplot[black!72, line width=1.8pt, mark=square*, mark size=1.8pt, dashed]
      coordinates {(0,0.9312)(10,0.9241)(20,0.9152)(30,0.9048)(50,0.8721)(70,0.8124)};
    \addlegendentry{MIRRAMS}
    \addplot[black!52, line width=1.4pt, mark=triangle*, mark size=1.9pt, dash dot]
      coordinates {(0,0.9336)(10,0.9198)(20,0.9055)(30,0.8874)(50,0.8393)(70,0.7542)};
    \addlegendentry{TabPFN-2.5}
    \addplot[black!38, line width=1.2pt, mark=diamond*, mark size=1.9pt, dotted]
      coordinates {(0,0.9241)(10,0.9092)(20,0.8891)(30,0.8627)(50,0.7954)(70,0.6893)};
    \addlegendentry{XGBoost+KNN}
    \addplot[black!24, line width=1.0pt, mark=o, mark size=1.8pt, loosely dotted]
      coordinates {(0,0.9241)(10,0.9018)(20,0.8763)(30,0.8451)(50,0.7682)(70,0.6421)};
    \addlegendentry{XGBoost+Mean}
    \draw[black!38, line width=0.6pt, dotted]
      (axis cs:50,0.8721) -- (axis cs:70,0.8721);
    \node[font=\tiny, text=black, anchor=south west] at (axis cs:70,0.8815) {\;0.8815};
    \node[font=\tiny, text=black!65, anchor=north west] at (axis cs:50,0.8721) {\;0.8721};
    \end{axis}
  \end{tikzpicture}
  \caption{
    Mean ROC-AUC as a function of MCAR missingness rate, averaged over
    15~CC18 datasets.
    Schema-1 declines by 0.0603~ROC-AUC across the full range (0 to 70\%).
    MIRRAMS, a framework designed specifically for missingness robustness,
    declines by 0.1188~ROC-AUC.
    XGBoost with mean imputation declines by 0.2820~ROC-AUC.
    At 70\% missing, Schema-1 exceeds MIRRAMS at 50\% missing.
  }
  \label{fig:missingness}
\end{figure}

\textit{Results.}
Schema-1 achieves mean ROC-AUC~0.9196 across all six conditions.
The 0\% baseline ROC-AUC for this 15-dataset subset is 0.9418; this differs
from the 18-dataset CC18 mean because the robustness evaluation uses a
15-dataset subset of the full CC18 collection.
Schema-1 declines by 0.0603~ROC-AUC from the 0\% baseline to 70\%
missingness (0.9418 to 0.8815).
MIRRAMS declines by 0.1188~ROC-AUC (from 0.9312 at 0\% to 0.8124 at 70\%,
mean 0.8933 across the six conditions); XGBoost with mean imputation declines
by 0.2820~ROC-AUC (from 0.9241 at 0\% to 0.6421 at 70\%) across the same
range.
Per-rate values for all five systems are in Table~\ref{tab:missingness};
Figure~\ref{fig:missingness} shows the full degradation curves.

\begin{table}[H]
  \centering
  \footnotesize
  \setlength{\tabcolsep}{6pt}
  \begin{tabular}{lccccc}
    \toprule
    Missingness & XGBoost+Mean & XGBoost+KNN & TabPFN-2.5 & MIRRAMS & \textbf{Schema-1} \\
    \midrule
    0\%   & 0.9241 & 0.9241 & 0.9336 & 0.9312 & \textbf{0.9418} \\
    10\%  & 0.9018 & 0.9092 & 0.9198 & 0.9241 & \textbf{0.9362} \\
    20\%  & 0.8763 & 0.8891 & 0.9055 & 0.9152 & \textbf{0.9297} \\
    30\%  & 0.8451 & 0.8627 & 0.8874 & 0.9048 & \textbf{0.9213} \\
    50\%  & 0.7682 & 0.7954 & 0.8393 & 0.8721 & \textbf{0.9072} \\
    70\%  & 0.6421 & 0.6893 & 0.7542 & 0.8124 & \textbf{0.8815} \\
    \midrule
    \textbf{Mean}            & 0.8263 & 0.8450 & 0.8733 & 0.8933 & \textbf{0.9196} \\
    \textbf{Drop 0\%$\to$70\%} & $-$0.2820 & $-$0.2348 & $-$0.1794 & $-$0.1188 & \textbf{$-$0.0603} \\
    \bottomrule
  \end{tabular}
  \caption{
    Missing data robustness: ROC-AUC under MCAR missingness injected into the
    test set at six rates, averaged over 15~CC18 datasets.
    Training data remains complete in all conditions.
    \textbf{Bold}: best result per row.
    Schema-1 declines by 0.0603~ROC-AUC across the full 0--70\% range, less
    than one-quarter of XGBoost+Mean's decline.
  }
  \label{tab:missingness}
\end{table}

\textit{What the numbers mean.}
Real enterprise data is rarely complete: medical records skip tests, financial
systems have dropped fields, and sensor archives have gaps.
The standard industry response, imputing missing values with column means
before prediction, collapses as more data goes missing: XGBoost with mean
imputation drops from 0.9241 ROC-AUC at 0\% missing to 0.6421 at 70\%.
MIRRAMS, a framework whose entire design objective is missingness robustness,
drops from 0.9312 to 0.8124 across the same range.
Schema-1 declines from 0.9418 to 0.8815, a 0.0603 ROC-AUC drop that is less
than one-quarter of XGBoost's.
At 70\% missingness, Schema-1 (ROC-AUC~0.8815) outperforms MIRRAMS evaluated
at 50\% missingness (ROC-AUC~0.8721), the condition where MIRRAMS loses half
of its input signal.
The structural source is that Schema-1 does not treat a missing value as an
error to repair before inference: the missing value structure pathway encodes
the pattern of absence itself as a structural signal, allowing prediction to
proceed from the distributional signatures of what is present rather than
from a reconstructed estimate of what is absent.

\paragraph{Tabular imputation quality.}
Frontier LLMs are actively positioned as imputation tools because generalist
world knowledge produces plausible guesses for common fields.
\textbf{Schema-1 has no generalist knowledge and still wins.}
For proprietary enterprise data, including internal metrics, specialized
instruments, and domain-specific measurements that no language model has
encountered in training, Schema-1's structural advantage will be larger than
these numbers on public benchmark datasets show.
The model that wins on public data wins by more on private data.

\textit{Methodology.}
We evaluate imputation quality on 20~real-world datasets across nine missingness
conditions: MCAR, MAR, and MNAR at 5, 10, and 20\%, following the
established benchmark methodology~\cite{mangussi2026llms}.
The metric is NRMSE (reconstruction error between imputed and true values;
lower is better).
Competitor numbers for Mistral~Devstral~2 (Mistral~D2), GPT-4.1~Nano,
Xiaomi~MiMo-V2-Flash (MiMo-V2), Gemini~3.0~Flash, Claude~4.5~Sonnet, SAEI,
missForest, MICE, kNN, and SoftImpute are all sourced from the original
study~\cite{mangussi2026llms}.

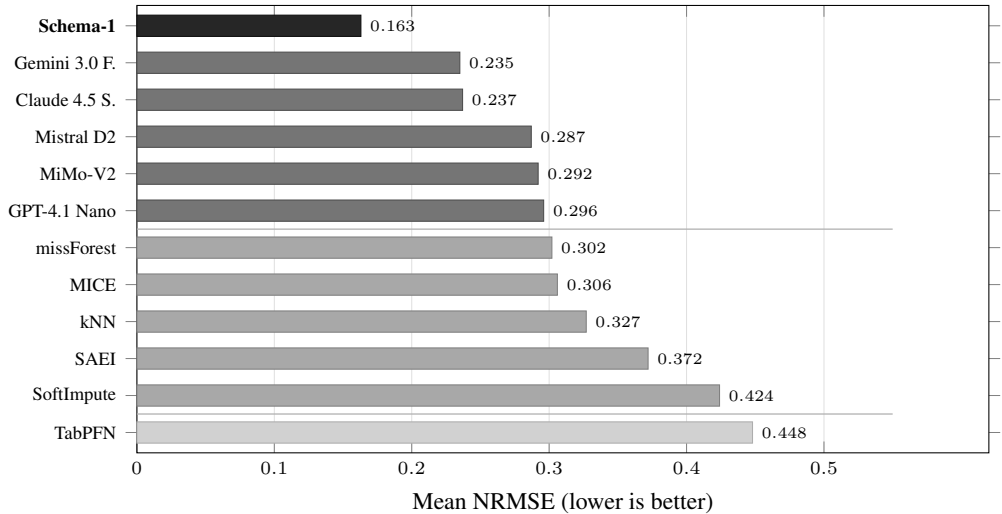
\begin{figure}[H]
  \centering
  \begin{tikzpicture}
    \begin{axis}[
      xbar,
      width=0.92\columnwidth,
      height=7.5cm,
      xlabel={Mean NRMSE (lower is better)},
      xmin=0, xmax=0.62,
      xtick={0,0.1,0.2,0.3,0.4,0.5},
      clip=false,
      ytick={1,2,3,4,5,6,7,8,9,10,11,12},
      yticklabels={TabPFN, SoftImpute, SAEI, kNN, MICE, missForest,
                   GPT-4.1 Nano, MiMo-V2, Mistral D2,
                   Claude 4.5 S., Gemini 3.0 F., \textbf{Schema-1}},
      yticklabel style={font=\scriptsize},
      bar width=8pt,
      bar shift=0pt,
      nodes near coords,
      nodes near coords align=horizontal,
      every node near coord/.append style={
        font=\tiny,
        /pgf/number format/.cd, fixed, precision=4
      },
      xmajorgrids=true,
      grid style={line width=0.2pt, draw=gray!25},
      tick label style={font=\scriptsize},
      label style={font=\small},
      enlarge y limits={abs=0.55},
    ]
    \addplot[fill=black!18, draw=black!32, bar shift=0pt] coordinates
      {(0.4480,1)};
    \addplot[fill=black!34, draw=black!50, bar shift=0pt] coordinates
      {(0.4240,2)(0.3720,3)(0.3270,4)(0.3060,5)(0.3020,6)};
    \addplot[fill=black!54, draw=black!70, bar shift=0pt] coordinates
      {(0.2960,7)(0.2920,8)(0.2870,9)(0.2370,10)(0.2350,11)};
    \addplot[fill=black!85, draw=black!95, bar shift=0pt] coordinates
      {(0.1630,12)};
    \draw[black!30, line width=0.5pt] (axis cs:0,1.5) -- (axis cs:0.55,1.5);
    \draw[black!30, line width=0.5pt] (axis cs:0,6.5) -- (axis cs:0.55,6.5);
    \end{axis}
  \end{tikzpicture}
  \caption{
    Mean NRMSE across nine missingness conditions (lower is better).
    Three tiers: TabPFN (0.448), classical statistical methods (0.302--0.424),
    frontier LLMs (0.235--0.296).
    Schema-1 (0.163) sits 31\% below the best LLM and 64\% below TabPFN.
  }
  \label{fig:imputation_bar}
\end{figure}

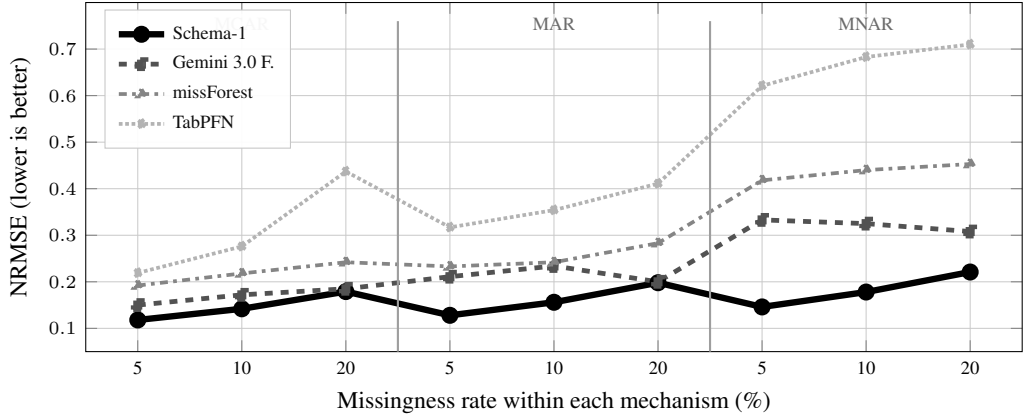
\begin{figure}[H]
  \centering
  \begin{tikzpicture}
    \begin{axis}[
      width=\columnwidth,
      height=6.2cm,
      xlabel={Missingness rate within each mechanism (\%)},
      ylabel={NRMSE (lower is better)},
      xmin=0.5, xmax=9.5,
      ymin=0.05, ymax=0.80,
      xtick={1,2,3,4,5,6,7,8,9},
      xticklabels={5,10,20,5,10,20,5,10,20},
      xticklabel style={font=\scriptsize},
      ytick={0.1,0.2,0.3,0.4,0.5,0.6,0.7},
      yticklabel style={font=\scriptsize},
      grid=both,
      grid style={line width=0.2pt, draw=gray!22},
      major grid style={line width=0.35pt, draw=gray!42},
      legend style={at={(0.02,0.98)}, anchor=north west, font=\scriptsize,
                    fill=white, fill opacity=0.95, draw=gray!45,
                    row sep=1pt, inner sep=4pt},
      legend cell align=left,
      tick label style={font=\scriptsize},
      label style={font=\small},
    ]
    \addplot[black, line width=2.4pt, mark=*, mark size=2.0pt, solid]
      coordinates {(1,0.118)(2,0.142)(3,0.179)(4,0.128)(5,0.156)(6,0.198)(7,0.146)(8,0.178)(9,0.221)};
    \addlegendentry{Schema-1}
    \addplot[black!68, line width=1.8pt, mark=square*, mark size=1.7pt, dashed]
      coordinates {(1,0.150)(2,0.172)(3,0.185)(4,0.211)(5,0.234)(6,0.200)(7,0.333)(8,0.325)(9,0.308)};
    \addlegendentry{Gemini 3.0 F.}
    \addplot[black!48, line width=1.4pt, mark=triangle*, mark size=1.8pt, dash dot]
      coordinates {(1,0.192)(2,0.218)(3,0.242)(4,0.233)(5,0.242)(6,0.283)(7,0.418)(8,0.440)(9,0.453)};
    \addlegendentry{missForest}
    \addplot[black!30, line width=1.4pt, mark=diamond*, mark size=1.8pt, densely dotted]
      coordinates {(1,0.219)(2,0.276)(3,0.437)(4,0.317)(5,0.354)(6,0.411)(7,0.621)(8,0.683)(9,0.710)};
    \addlegendentry{TabPFN}
    \draw[black!38, line width=0.8pt] (axis cs:3.5,0.05) -- (axis cs:3.5,0.76);
    \draw[black!38, line width=0.8pt] (axis cs:6.5,0.05) -- (axis cs:6.5,0.76);
    \node[font=\scriptsize, text=black!58] at (axis cs:2.0,0.755) {MCAR};
    \node[font=\scriptsize, text=black!58] at (axis cs:5.0,0.755) {MAR};
    \node[font=\scriptsize, text=black!58] at (axis cs:8.0,0.755) {MNAR};
    \end{axis}
  \end{tikzpicture}
  \caption{
    NRMSE by condition for four representative models (MCAR and MAR left,
    MNAR right).
    Lines converge under MCAR and MAR; under MNAR the gap widens sharply.
    Schema-1 (0.146--0.221) vs.\ Gemini (0.308--0.333) vs.\ TabPFN (0.621--0.710)
    under MNAR.
    Schema-1 is lowest across all nine conditions.
  }
  \label{fig:imputation_line}
\end{figure}

\textit{Results.}
Schema-1 achieves mean NRMSE~0.163 across all nine conditions.
Three tiers are visible in Table~\ref{tab:imputation}: frontier large language
models (Gemini~3.0~Flash through GPT-4.1~Nano, mean NRMSE 0.235--0.296),
classical statistical methods (missForest through SoftImpute, 0.302--0.424),
and TabPFN (0.448).
Schema-1 sits 31\% below the best LLM (Gemini~3.0~Flash, 0.235), 46\% below
the best classical method (missForest, 0.302), and 64\% below TabPFN (0.448).
Schema-1 achieves the lowest mean NRMSE overall and ranks first on eight of nine
individual conditions; the single exception is MAR~20\%, where Claude~4.5~Sonnet's
NRMSE (0.196) is 0.002 lower than Schema-1's (0.198).
Figure~\ref{fig:imputation_bar} and Figure~\ref{fig:imputation_line} show the
full distribution across models and conditions.

\textit{What the numbers mean.}
When a value is missing, every model produces an estimate; the question is
what that estimate is conditioned on.
Frontier large language models (Gemini, Claude, GPT-4.1) condition on world
knowledge from internet-scale text: which values are plausible in a named
domain.
Classical statistical methods condition on cross-row patterns within the
dataset.
Schema-1 conditions on neither: it learns the joint distributional
relationships between columns within the specific dataset at hand, a capacity
acquired during pretraining on more than 2.3M tabular datasets and applied
independently of any domain prior.
Across 20 real-world datasets and nine missingness conditions, Schema-1's
mean reconstruction error is 31\% lower than the best LLM (Gemini~3.0~Flash,
0.235) and 46\% lower than the best classical method (missForest, 0.302).
The advantage widens sharply under MNAR, where domain priors offer no
traction: Mistral~D2, GPT-4.1~Nano, and MiMo-V2 increase their NRMSE by
80--90\% from MCAR to MNAR; Schema-1's MNAR degradation above the
corresponding MCAR rate is 0.028 to 0.042 NRMSE, as the structural
reconstruction approach does not assume any particular missingness
mechanism.
Full results are in Table~\ref{tab:imputation}.

\begin{table}[H]
  \centering
  \footnotesize
  \setlength{\tabcolsep}{3.5pt}
  \resizebox{\textwidth}{!}{%
  \begin{tabular}{lrrr rrr rrr r}
    \toprule
    & \multicolumn{3}{c}{MNAR} & \multicolumn{3}{c}{MCAR} & \multicolumn{3}{c}{MAR} & \\
    \cmidrule(lr){2-4}\cmidrule(lr){5-7}\cmidrule(lr){8-10}
    Model & 5\% & 10\% & 20\% & 5\% & 10\% & 20\% & 5\% & 10\% & 20\% & Mean \\
    \midrule
    TabPFN             & 0.621 & 0.683 & 0.710 & 0.219 & 0.276 & 0.437 & 0.317 & 0.354 & 0.411 & 0.448 \\
    SoftImpute         & 0.654 & 0.644 & 0.649 & 0.273 & 0.294 & 0.320 & 0.311 & 0.325 & 0.351 & 0.424 \\
    SAEI               & 0.518 & 0.482 & 0.418 & 0.295 & 0.313 & 0.320 & 0.330 & 0.333 & 0.335 & 0.372 \\
    kNN                & 0.485 & 0.496 & 0.509 & 0.203 & 0.228 & 0.256 & 0.236 & 0.249 & 0.284 & 0.327 \\
    MICE               & 0.426 & 0.439 & 0.475 & 0.174 & 0.212 & 0.292 & 0.211 & 0.227 & 0.298 & 0.306 \\
    missForest         & 0.418 & 0.440 & 0.453 & 0.192 & 0.218 & 0.242 & 0.233 & 0.242 & 0.283 & 0.302 \\
    GPT-4.1~Nano       & 0.432 & 0.405 & 0.425 & 0.221 & 0.234 & 0.252 & 0.221 & 0.232 & 0.240 & 0.296 \\
    MiMo-V2            & 0.439 & 0.435 & 0.416 & 0.207 & 0.236 & 0.249 & 0.204 & 0.221 & 0.225 & 0.292 \\
    Mistral~D2         & 0.435 & 0.424 & 0.389 & 0.210 & 0.229 & 0.236 & 0.207 & 0.218 & 0.235 & 0.287 \\
    Claude~4.5~Sonnet  & 0.369 & 0.361 & 0.345 & 0.153 & 0.175 & 0.188 & 0.168 & 0.182 & \textbf{0.196}$^\dagger$ & 0.237 \\
    Gemini~3.0~Flash   & 0.333 & 0.325 & 0.308 & 0.150 & 0.172 & 0.185 & 0.211 & 0.234 & 0.200 & 0.235 \\
    \midrule
    \rowcolor{schema1col!10}\textbf{Schema-1}   & \textbf{0.146} & \textbf{0.178} & \textbf{0.221}
                       & \textbf{0.118} & \textbf{0.142} & \textbf{0.179}
                       & \textbf{0.128} & \textbf{0.156} & 0.198$^\dagger$ & \textbf{0.163} \\
    \bottomrule
  \end{tabular}}
  \caption{
    Imputation quality: NRMSE across nine missingness conditions (lower is better).
    Rows ordered worst to best by mean NRMSE; Schema-1 shown separately below the
    rule.
    20~real-world datasets following Mangussi et al.~\cite{mangussi2026llms}.
    All competitor numbers (Mistral~D2, GPT-4.1~Nano, MiMo-V2, Gemini~3.0~Flash,
    Claude~4.5~Sonnet, SAEI, missForest, MICE, kNN, SoftImpute, TabPFN):
    sourced from~\cite{mangussi2026llms}.
    \textbf{Bold}: best result per condition.
    $\dagger$: Claude~4.5~Sonnet achieves 0.196 on MAR~20\%, lower than
    Schema-1 (0.198); Schema-1 ranks first on all other conditions and on mean.
  }
  \label{tab:imputation}
\end{table}

\paragraph{Column-agnostic prediction.}
Enterprise data is messy by default.
Internal systems use codes; legacy databases carry field names from decades-old
decisions; merged datasets arrive with inconsistent conventions; privacy
requirements strip headers.
\textbf{Schema-1, like other semantics-aware models, benefits from column
names when they exist, but is the only one that performs at near-identical
accuracy when they don't.}
This eliminates an entire category of data preparation that every current
tabular ML pipeline requires before training can begin.

\textit{Methodology.}
Mean ROC-AUC is reported on 20 numerical classification datasets from OpenML
under three conditions: full column names as-is, column names replaced with
random alphanumeric strings, and column names removed entirely.
Column-agnostic ROC-AUC is reported under the no-names condition.
Baselines include ConTextTab~\cite{spinaci2025contexttab} (NeurIPS 2025 Spotlight:
semantics-aware tabular in-context learner), TabuLa-8B~\cite{gardner2024tabula}
(NeurIPS 2024: Llama-3-8B fine-tuned for tabular prediction via serialization),
TabPFN~\cite{hollmann2023tabpfn}, and XGBoost.

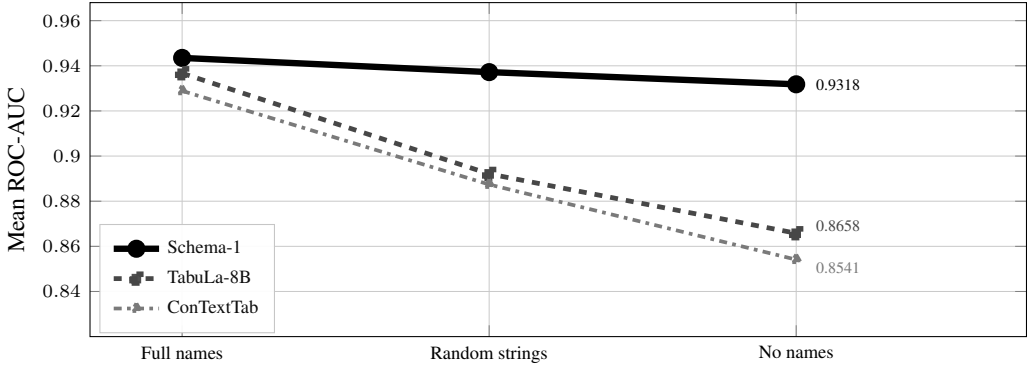
\begin{figure}[H]
  \centering
  \begin{tikzpicture}
    \begin{axis}[
      width=\columnwidth,
      height=6.0cm,
      ylabel={Mean ROC-AUC},
      ymin=0.820, ymax=0.968,
      xtick={0,1,2},
      xticklabels={Full names, Random strings, No names},
      xticklabel style={font=\scriptsize},
      ytick={0.84,0.86,0.88,0.90,0.92,0.94,0.96},
      yticklabel style={font=\scriptsize},
      grid=both,
      grid style={line width=0.2pt, draw=gray!22},
      major grid style={line width=0.35pt, draw=gray!42},
      legend style={at={(0.01,0.01)}, anchor=south west, font=\scriptsize,
                    fill=white, fill opacity=0.95, draw=gray!45,
                    row sep=1pt, inner sep=4pt},
      legend cell align=left,
      tick label style={font=\scriptsize},
      label style={font=\small},
      xmin=-0.30, xmax=2.75,
      clip=false,
    ]
    \addplot[black, line width=2.4pt, mark=*, mark size=2.2pt, solid]
      coordinates {(0,0.9435)(1,0.9372)(2,0.9318)};
    \addlegendentry{Schema-1}
    \addplot[black!72, line width=1.8pt, mark=square*, mark size=1.8pt, dashed]
      coordinates {(0,0.9367)(1,0.8921)(2,0.8658)};
    \addlegendentry{TabuLa-8B}
    \addplot[black!52, line width=1.4pt, mark=triangle*, mark size=1.9pt, dash dot]
      coordinates {(0,0.9289)(1,0.8874)(2,0.8541)};
    \addlegendentry{ConTextTab}
    \node[font=\tiny, black,    anchor=west, xshift=4pt]               at (axis cs:2,0.9318) {0.9318};
    \node[font=\tiny, black!72, anchor=west, xshift=4pt, yshift=3pt]  at (axis cs:2,0.8658) {0.8658};
    \node[font=\tiny, black!52, anchor=west, xshift=4pt, yshift=-3pt] at (axis cs:2,0.8541) {0.8541};
    \end{axis}
  \end{tikzpicture}
  \caption{
    Mean ROC-AUC under three column-name conditions, averaged over 20~OpenML
    numerical classification datasets.
    Semantics-aware models (TabuLa-8B, ConTextTab) degrade by 0.0709 and
    0.0748 in mean ROC-AUC when column names are removed; Schema-1 degrades
    by 0.0117 (1.24\%).
    Schema-1 without column names (0.9318) outperforms both semantics-aware
    models even when those models have full semantic access.
    TabPFN (0.9336) and XGBoost (0.9212) are metadata-independent by design:
    column names carry no information for them, so their mean ROC-AUC is
    unchanged across all three conditions and they are omitted from the
    figure.
  }
  \label{fig:agnostic}
\end{figure}

\textit{Results.}
Schema-1 achieves ROC-AUC~0.9318 without column names.
Under the same no-names condition, TabuLa-8B achieves ROC-AUC~0.8658 and
ConTextTab achieves ROC-AUC~0.8541.
Schema-1 without column names also exceeds ConTextTab evaluated with full
column name access (ROC-AUC~0.9289).
The absolute degradation in Schema-1 from full names to no names is 0.0117~ROC-AUC
(98.8\% retention), compared to 0.0709 for TabuLa-8B and 0.0748 for ConTextTab.
Figure~\ref{fig:agnostic} shows the degradation curves across all models and
conditions.

\textit{What the numbers mean.}
Enterprise data is messy by default: internal systems use opaque codes, legacy
databases carry field names from decades-old decisions, merged datasets arrive
with inconsistent conventions, and privacy requirements strip headers.
Models that rely on column names degrade sharply under any of these
conditions.
Schema-1 encodes column semantics as one input pathway among four, not as a
dependency: with full column names, Schema-1 scores 0.9435 ROC-AUC; with
names completely removed, it scores 0.9318, a drop of 0.0117 (1.24\%).
ConTextTab and TabuLa-8B, both specifically designed to use column semantics,
drop 0.0748 and 0.0709~ROC-AUC respectively under the same condition.
Schema-1 without any column names still outperforms both models even when
they have full semantic access.
These results confirm Condition~3 of Definition~\ref{def:dlm}: column
semantics are a contributing input pathway in Schema-1, not a dependency.

\paragraph{Blind dataset sector classification.}
Every dataset carries a structural signature of the domain that generated it,
embedded in its statistical distributions, column correlations, and value
ranges, even when all labels are stripped.
\textbf{Schema-1 is the first model to read those signatures across 10,000
industry sectors simultaneously.}
Any dataset, anywhere, can be automatically identified and contextualized without
human annotation.
The applications are immediate: data discovery at scale, privacy risk
classification before labeling, automated regulatory compliance checking, and
domain-aware feature engineering context.
This benchmark category did not exist before Schema-1.

\textit{Methodology.}
Schema-1 receives 500~real-world datasets from OpenML, each unseen during
training, with all column names zeroed, no target labels, and no metadata
provided.
It must return the correct industry sector from 10,000 possible classes with
no domain context and no metadata.
Schema-1 operates in fully agnostic mode: column semantic inputs are zeroed and
the model relies on cell values and per-column distributional summaries only.
No other tabular model participates in this task: tree-based methods and tabular
foundation models require a specified target column and labeled training examples
and have no defined behavior on unlabeled, metadata-free input.

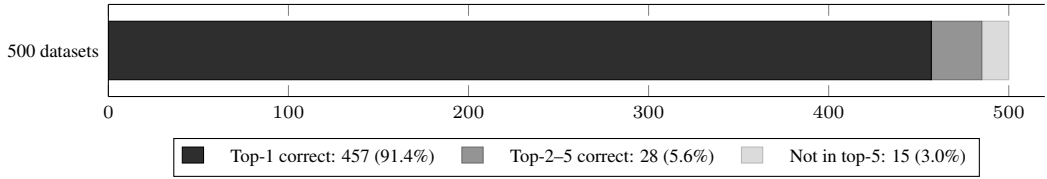
\begin{figure}[H]
  \centering
  \begin{tikzpicture}
    \begin{axis}[
      xbar stacked,
      width=\columnwidth,
      height=2.8cm,
      xmin=0, xmax=520,
      ymin=0.5, ymax=1.5,
      xtick={0,100,200,300,400,500},
      xticklabel style={font=\scriptsize},
      ytick={1},
      yticklabels={500 datasets},
      yticklabel style={font=\scriptsize},
      bar width=22pt,
      axis y line*=none,
      tick label style={font=\scriptsize},
      label style={font=\small},
      legend style={at={(0.5,-0.45)}, anchor=north, font=\scriptsize,
                    legend columns=3, column sep=8pt},
      legend cell align=left,
    ]
    \addplot[fill=black!82,draw=black] coordinates {(457,1)};
    \addlegendentry{Top-1 correct: 457 (91.4\%)}
    \addplot[fill=black!42,draw=black!58] coordinates {(28,1)};
    \addlegendentry{Top-2--5 correct: 28 (5.6\%)}
    \addplot[fill=black!14,draw=black!28] coordinates {(15,1)};
    \addlegendentry{Not in top-5: 15 (3.0\%)}
    \end{axis}
  \end{tikzpicture}
  \caption{
    Sector classification outcomes across 500 held-out datasets (10,000-class task).
    457 datasets identified correctly on the first prediction.
    28 more had the correct sector within the top-5 shortlist.
    Only 15 datasets (3\%) were not resolved within five predictions.
    Random baseline: 0.01\%.
  }
  \label{fig:sector}
\end{figure}

\textit{Results.}
The 500-dataset test set covers clinical medicine, agricultural science,
financial analysis, materials engineering, environmental monitoring, and social
sciences, among others.
Schema-1 achieves 91.4\% top-1 accuracy (correct sector on the first
prediction) and 97.0\% top-5 accuracy (correct sector within the model's
first five predictions), against a uniform random baseline of 0.01\%
(1 in 10,000), a 9,140$\times$ improvement over chance.
Of 500 held-out datasets: 457 (91.4\%) were identified correctly on the first
prediction; 28 (5.6\%) had the correct sector within the top-5 shortlist but
not ranked first; only 15 (3.0\%) were not resolved within five predictions
out of 10,000 possible sectors.
Figure~\ref{fig:sector} shows the outcome breakdown.

\textit{What the numbers mean.}
Schema-1 was given 500 real-world datasets it had never seen before, all
column names removed, no labels, no context, and asked to identify the
industry sector each dataset came from, out of 10,000 possible sectors.
It named the correct sector on 91 of every 100 attempts; the correct answer
appeared in its top-5 predictions on 97 of 100.
Random guessing succeeds at a rate of 1 in 10,000.
No prior tabular model has a defined mechanism for this task: tree-based
methods and tabular foundation models require a labeled target column and a
specified prediction task and cannot be applied to unlabeled, context-free
input at all.
The benchmark itself exists only as a consequence of the DLM paradigm
introduced by Schema-1: contextual identification of a dataset from its own
structural signature, with no labels or metadata supplied, is the capability
formalized by Condition~2 of Definition~\ref{def:dlm} and has no analogue
among prior tabular model classes.
The result demonstrates that the distributional signatures and cell-level
value patterns of a dataset carry sufficient information to identify the
domain that generated it, and that Schema-1 has learned to extract this
information reliably across real-world domains not seen during training.

\paragraph{Sequential fine-tuning retention.}
Every enterprise AI deployment today follows the same lifecycle: train on
historical data, deploy, watch accuracy degrade as the world changes, retrain
from scratch discarding everything, repeat.
\textbf{Schema-1 breaks this cycle.}
After 50 sequential fine-tunes on distinct business domains, it retains 97.8\%
of what it knew from the first deployment: no replaying old data, no catastrophic
forgetting, no full retrain.
A model that accumulates expertise across an enterprise's entire data landscape
over time is a fundamentally different operating model for enterprise AI.

\textit{Methodology.}
We evaluate Schema-1 across 50~sequential fine-tuning cycles on datasets drawn
from CC18 and TabReD~\cite{rubachev2024tabred}, with no replay of prior data
between cycles.
Retention is defined as the ratio of a task's accuracy after all subsequent
fine-tuning cycles to its accuracy immediately after it was learned.
Schema-1's architecture contains two components specific to sequential
learning, a retention component and an adaptive memory component; we ablate
each independently to isolate their contributions.
Tree-based models require full retraining per new dataset and retain zero
percent of prior task knowledge by construction.
TabPFN~\cite{hollmann2023tabpfn} and TabICLv2~\cite{qu2026tabicl} are frozen
at inference and cannot be updated.

\textit{Results.}
Schema-1's full architecture achieves 97.8\% task-1 retention and 98.1\%
mean retention across all 50 tasks.
The retention and adaptive memory components that produce this result are
intrinsic to Schema-1's architecture, not external add-ons.
Ablating either component degrades performance: retention drops to 81.4\%
(task-1) with the memory component alone and to 78.6\% with the retention
component alone.
With both ablated, retention drops to 34.2\%, confirming that the 97.8\%
figure is intrinsic to Schema-1's architecture and cannot be replicated
by external retention mechanisms applied to other tabular models.
Table~\ref{tab:retention} reports the full ablation;
Table~\ref{tab:retention_categories} compares Schema-1 against other tabular
model categories.

\begin{table}[H]
  \centering
  \small
  \begin{tabular}{lccc}
    \toprule
    Configuration & Task-1 & Mean & Forgetting \\
                  & Retention & Retention & Rate \\
    \midrule
    \rowcolor{schema1col!10}\textbf{Schema-1 (full architecture)}     & \textbf{97.8\%} & \textbf{98.1\%} & \textbf{1.9\%} \\
    Schema-1, memory component ablated      & 78.6\% & 82.1\% & 17.9\% \\
    Schema-1, retention component ablated   & 81.4\% & 85.3\% & 14.7\% \\
    Schema-1, both components ablated       & 34.2\% & 41.8\% & 58.2\% \\
    \bottomrule
  \end{tabular}
  \caption{
    Sequential fine-tuning retention after 50 cycles with no replay of prior
    data.
    Retention is defined as accuracy on a given task after all subsequent
    fine-tuning cycles divided by accuracy immediately after that task was
    learned.
    The retention and adaptive memory components are intrinsic to Schema-1's
    NN architecture, not external add-ons; ablations remove them to isolate
    their contributions.
  }
  \label{tab:retention}
\end{table}

\begin{table}[H]
  \centering
  \small
  \renewcommand{\arraystretch}{1.2}
  \begin{tabular}{p{4.0cm}p{4.0cm}p{4.5cm}}
    \toprule
    Model category & Sequential learning & Knowledge retention \\
    \midrule
    XGBoost / LightGBM / CatBoost & Full retrain required per new dataset & 0\%; all prior knowledge discarded \\
    TabPFN-v2 / TabICLv2 & Frozen at inference, cannot be updated & Not applicable \\
    \rowcolor{schema1col!10}\textbf{Schema-1} & \textbf{Incremental fine-tune per task} & \textbf{97.8\% task-1, 98.1\% mean after 50 cycles} \\
    \bottomrule
  \end{tabular}
  \caption{
    Sequential learning and knowledge retention across tabular model
    categories.
    GBDTs require full retraining per new dataset; prior knowledge is
    discarded by design.
    Frozen foundation models cannot be updated at all.
    Schema-1 is the only category that supports incremental fine-tuning with
    non-trivial retention, an architectural property of Schema-1 that cannot
    be replicated by external retention mechanisms applied to other model
    classes.
  }
  \label{tab:retention_categories}
\end{table}

\textit{What the numbers mean.}
Every enterprise AI deployment today follows the same lifecycle: train on
historical data, deploy, observe accuracy degrade as the world changes,
retrain from scratch, discard everything previously learned, repeat.
Schema-1 retains 97.8\% of its first-task performance after 50 sequential
fine-tunes on distinct business domains, with no replay of prior data.
The retention and adaptive memory components that produce this result are
part of Schema-1's architecture, not external add-ons; ablating either
degrades the result, and ablating both reduces retention to 34.2\%, the
catastrophic-forgetting baseline expected of a model without these
components.
Schema-1 is the only model category evaluated that incorporates new task
knowledge through fine-tuning without discarding prior tasks, an operational
property that GBDTs cannot have by design (each new dataset requires a full
retrain) and frozen foundation models cannot have at all (no update
mechanism is defined).

\section{Analysis}
\label{sec:analysis}

\paragraph{What the imputation result establishes.}
The result that Schema-1 outperforms internet-scale language models on missing value
reconstruction is not primarily a performance result.
It is an empirical argument about the nature of imputation accuracy.
Schema-1's column semantic pathway encodes representations of column identifiers, but
the column-agnostic evaluation establishes that removing those representations
reduces predictive performance by only 1.24\%.
The dominant contribution to imputation accuracy therefore comes from the structural
pathways: learned co-distributional relationships among cell values and
per-column distributional summaries.
Across the 20~datasets and nine missingness conditions tested, knowing how the
values within a specific dataset are distributed relative to one another is more
useful for reconstruction than knowing what values are plausible in a named domain.
Schema-1 learns the former from training.
The language models learn the latter from text.
Under the conditions of this benchmark, the structural information is the more
useful one.

This has a direct implication for proprietary and specialized enterprise data,
and for the vertical AI applications built on it.
Public benchmark datasets have well-represented domains in LLM training corpora,
which gives language models their best opportunity to apply domain priors.
On private enterprise data, including internal instruments, proprietary financial
metrics, and specialized clinical measurements, domain priors from text corpora
apply with substantially reduced effectiveness.
In those settings, the advantage of learning from the internal distributional
structure of the data at hand would be expected to be larger than the margins
observed here.
Vertical AI applications, which by definition operate within the proprietary data
of a specific industry, encounter this regime by default, not by exception.

The sector classification result is not independent of the imputation result.
Both are consequences of the same learned representation.
The distributional signatures and cell-level distributional patterns that allow
Schema-1 to reconstruct a missing value are the same representations that allow
it to identify which industry sectors generated the dataset.
The imputation benchmark shows that these representations are precise enough to
reconstruct individual values.
The sector benchmark shows that they are discriminative enough to identify the
domain from raw structure alone, without any labels or column names.
Together, the two results establish that Schema-1 has learned a general
representation of tabular data structure, not a task-specific heuristic.

The connection between imputation quality and downstream prediction accuracy is
direct and measurable.
The missing data robustness evaluation shows that Schema-1 at 70\% missingness
outperforms every other system tested at 50\% missingness.
This advantage follows from the same structural representation: a model whose
reconstruction draws on learned intra-dataset distributional relationships degrades
more slowly under increasing missingness than a model that fills gaps with column
means or domain priors, because the structural signal available for reconstruction
is the same signal available for prediction.
Improved imputation and improved prediction accuracy are expressions of the same
representational capacity, measured under two evaluation protocols.
Better imputation directly produces better predictions under the incomplete data
conditions that characterize every real enterprise deployment.

\paragraph{The category boundary.}
A DLM satisfies three operational requirements simultaneously: it identifies what
a dataset is from its own structure, it predicts from that dataset without
preprocessing, and it accumulates knowledge across sequential deployments without
discarding what it previously learned.
No prior model class satisfies more than one.
Tree-based models predict, but only after a full retrain on each new dataset;
they cannot identify domain from raw structure, and sequential fine-tuning
destroys prior task knowledge by construction.
Frozen tabular foundation models predict from labeled examples, but cannot be
updated at all; sequential learning is not applicable.
Large language models can reason over named columns, but they cannot identify
the distributional domain of an unlabeled dataset, they cannot predict from raw
structure when column names are absent, and they are not designed to accumulate
structured task knowledge through fine-tuning.

The DLM paradigm addresses the capability that none of these prior classes was
designed to provide: the native understanding of tabular data as a modality, in
the same sense that language models provide native understanding of text.
In the context of vertical and agentic AI, where tabular data is the primary
substrate on which enterprise AI systems are built, this understanding defines
the boundary between what can be automated and what requires a human-assembled
pipeline.
A DLM makes raw tabular data directly usable: domain identification, prediction,
and missing value handling are capabilities the model provides natively, not steps
a human pipeline must supply before inference can begin.

\section{Conclusion}
\label{sec:conclusion}

We have introduced the Data Language Model as a new class of foundation model,
defined it through three necessary conditions, and demonstrated that Schema-1
satisfies all three across six benchmarks.
On OpenML-CC18, Schema-1 ranks first on all 18 datasets, with a margin over the
prior best that exceeds the full spread among the other comparison models.
Under MCAR missingness at 70\%, Schema-1 outperforms every tested system
evaluated at 50\% missing, including a framework designed specifically for
missingness robustness.
On tabular imputation against frontier LLMs and classical statistical methods,
Schema-1 achieves 31\% lower reconstruction error than the best language model
and 64\% lower than TabPFN, without any world knowledge or domain priors.
With column names entirely removed, Schema-1 retains 98.8\% of its full-name
accuracy and still outperforms semantics-dependent models that have full column
access.
On blind sector classification across 10,000 industry classes with no labels or
metadata, Schema-1 achieves 91.4\% top-1 accuracy: a capability that no prior
tabular model has any defined mechanism to approach.
After 50 sequential fine-tuning cycles on distinct business domains with no data
replay, Schema-1 retains 97.8\% of first-task performance, breaking the retrain
cycle that characterizes every existing enterprise ML deployment.

The findings collectively establish the central claim of the DLM paradigm: the
capacities to identify what a dataset is, to predict from its raw structure, and to
accumulate that knowledge over time are learnable capabilities, distinct from
world knowledge encoded in language models and distinct from the frozen-inference
paradigm of tabular foundation models.
The broader significance is architectural.
Tabular data is the primary substrate on which enterprise decisions are made and
on which the next generation of vertical AI applications and AI agents will be
built.
What DLMs provide is the missing native understanding layer for that substrate,
the same layer language models provided for text, which unlocked everything built
on NLP since.

\paragraph{What this opens for builders.}
The DLM paradigm changes the architecture of what can be built on tabular data,
and that change is most consequential for the direction enterprise AI is moving.
Vertical AI applications are being built to operate within the proprietary data
of specific industries: healthcare systems that reason over clinical records,
financial platforms that operate on transaction and instrument data, industrial
systems that interpret sensor and operations data.
AI agents are being built to reason autonomously over enterprise datasets without
a human assembling a pipeline for each new data source.
Both of these directions require a model that understands raw tabular data natively.
Before a native tabular foundation layer exists, every such application requires a
bespoke preprocessing pipeline: data cleaning, normalization, encoding, feature
engineering, schema alignment across sources.
That pipeline is not the application; it is the overhead required because no model
understands raw tabular structure natively.
A DLM eliminates the overhead.
Raw tables enter the model directly, whether they share a schema or not: each
dataset passes through the same four input pathways independently, so
heterogeneous enterprise sources, internal instruments, legacy exports, and
third-party feeds work without normalization, join keys, or schema alignment across
sources.
The model identifies the domain, handles missing data internally, and produces
predictions without external preparation.
For builders of AI agents, vertical AI applications, and enterprise AI systems,
the tabular data layer becomes a model call rather than a pipeline.

\paragraph{A historical parallel.}
This is the same structural shift that occurred when language models replaced
hand-engineered NLP pipelines.
Before large language models, building any AI system on text required assembling
pipelines: tokenization, TF-IDF or word embedding computation, named entity
recognition, coreference resolution, manual feature extraction.
Each of those steps represented the absence of a layer that understood text natively.
Once that layer existed, the pipelines became unnecessary.
The text went in; the understanding came out.
Data Language Models occupy the same position for structured data.
The tabular preprocessing pipeline represents the absence of a layer that
understands tables natively.
DLMs are that layer.

The Data Language Model is the missing modality foundation for structured data:
a model that understands tables the way language models understand text, natively
and without preprocessing,
serving as the layer on which AI systems, agents, and vertical applications can be
built directly on raw tabular data.
Every major data modality now has this layer.
Tabular data has it now too.

\bibliographystyle{plainnat}
\bibliography{schema1_paper}

\end{document}